\documentclass[a4paper,11pt]{article}

\newcommand{\model}{LogicVista~}
\newcommand{\modelNoSpace}{LogicVista}
\newcommand{\datacount}{448~}
\newcommand{\MLLMcount}{8~}
\newcommand{\reasoningcount}{5~}
\newcommand{\capcount}{9~}

\newcommand{\inductive}{107}
\newcommand{\deductive}{93}
\newcommand{\numerical}{95}
\newcommand{\spatial}{79}
\newcommand{\mech}{74}
\newcommand{\diagram}{330}
\newcommand{\ocr}{234}
\newcommand{\patternings}{105}
\newcommand{\graphs}{67}
\newcommand{\tables}{70}
\newcommand{\dshapes}{45}
\newcommand{\puzzles}{256}
\newcommand{\sequences}{76}
\newcommand{\physics}{69}

\newcommand\blfootnote[1]{%
  \begingroup
  \renewcommand\thefootnote{}\footnote{#1}%
  \addtocounter{footnote}{-1}%
  \endgroup
}

\usepackage[margin=1in]{geometry} 

\usepackage{times}
\usepackage{latexsym}

\usepackage[T1]{fontenc}
\usepackage[utf8]{inputenc}

\usepackage{microtype}

\usepackage{inconsolata}

\usepackage{amsmath}
\usepackage{amssymb}
\usepackage{amsthm}
\usepackage{amsfonts}
\usepackage{mathtools}

\usepackage{array}
\usepackage{calc}
\usepackage{siunitx}
\usepackage{makecell}
\usepackage{tabularx}

\usepackage[numbers,compress]{natbib}

\usepackage[colorlinks,citecolor=blue]{hyperref}
\usepackage{url}

\usepackage{booktabs}
\usepackage{graphicx}
\usepackage{caption}
\usepackage{colortbl}
\usepackage{float}

\usepackage{xcolor}
\usepackage{enumitem}
\usepackage{framed}
\usepackage{xspace}

\usepackage{multirow}
\usepackage{multicol}

\usepackage{pgfplots}
\usepackage{graphics}
\usepackage{arydshln}

\usepackage{tablefootnote}

\usepackage[toc,page,header]{appendix}
\usepackage{minitoc}

\usepackage{todonotes}
\usepackage{subcaption}
\usepackage{amssymb}
\usepackage{pifont}
\newcommand{\cmark}{\ding{51}}%
\newcommand{\xmark}{\ding{55}}%
\graphicspath{ {./figures} }

\title{LogicVista: Multimodal LLM Logical Reasoning Benchmark in Visual Contexts}

\author{
    Yijia Xiao$^{1*}$,
    Edward Sun$^{1*}$,
    Tianyu Liu$^{2}$,
    Wei Wang$^{1}$ \\
    $^{1}$University of California, Los Angeles, \\
    $^{2}$Yale University \\
    $^{1}$\texttt{\{yijia.xiao, weiwang\}@cs.ucla.edu}, \\
    $^{1}$\texttt{edwardsun12895@g.ucla.edu} \\
    $^{2}$\texttt{tianyu.liu@yale.edu}
}

\date{}

\begin{document}
\maketitle
\begin{abstract}
We propose \modelNoSpace, an evaluation benchmark that assesses the integrated logical reasoning capabilities of multimodal large language models (MLLMs) in \textbf{Vis}ual contexts. Recent advancements in MLLMs have demonstrated various fascinating abilities, from crafting poetry based on an image to performing mathematical reasoning. However, there is still a lack of systematic evaluation of MLLMs' proficiency in 
logical reasoning tasks, which are essential for activities like navigation and puzzle-solving. Thus we evaluate general logical cognition abilities across \reasoningcount logical reasoning tasks encompassing \capcount different capabilities, using a sample of \datacount multiple-choice questions. Each question is annotated with the correct answer and the human-written reasoning behind the selection, enabling both open-ended and multiple-choice evaluation. A total of \MLLMcount MLLMs are comprehensively evaluated using \modelNoSpace. Code and Data Available at \url{https://github.com/Yijia-Xiao/LogicVista}.
\end{abstract}

\blfootnote{$^\ast$ Both authors contributed equally.}

\section{Introduction} \label{intro}
Recent advancements in Large Language Models (LLMs) are gradually turning the vision of a generalist AI agent into reality. These models exhibit near-human expert-level performance across a variety of tasks and have recently been augmented with visual understanding capabilities, enabling them to tackle even more complex visual challenges. This branch of work, led by proprietary projects such as GPT-4 \citep{openai2024gpt4} and Flamingo \citep{alayrac2022flamingo}, as well as open-source efforts like LLaVA \citep{liu2023llava}, Mini-GPT4 \citep{zhu2023minigpt4}, enhances existing LLMs by incorporating visual comprehension. These models, known as Multimodal Large Language Models (MLLMs), use LLMs as the foundation for processing information and generating reasoned outcomes \citep{Yin}, thereby bridging the gap between language and vision. 

\begin{figure*}[tbp]
\begin{center}
\begin{tabular}{c c c c c c c c c c c} 
\textbf{\model (Ours)} & & & & & & & & & \textbf{VQAv2, TextVQA and MM-vet} 
\end{tabular}
\begin{tabular}{c c c c} 
    \\ 
    \raisebox{-.25\totalheight}{\includegraphics[width=2.65cm]{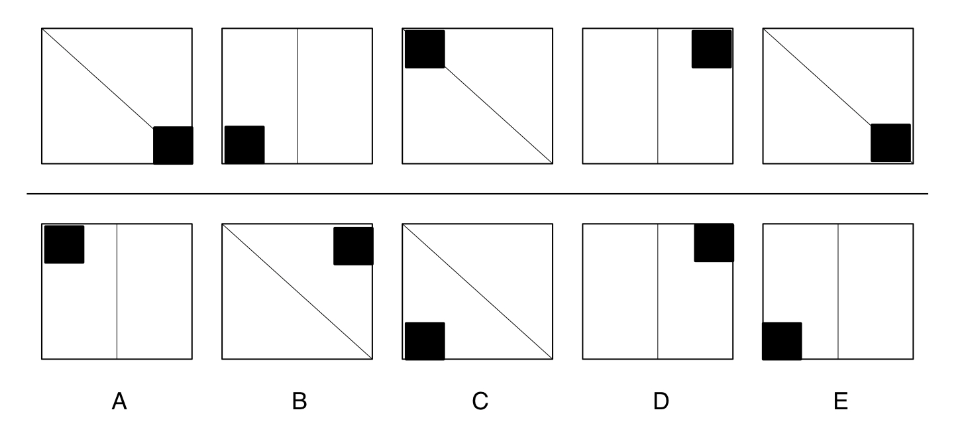}} & \begin{tabular}{c} 
    \tiny{\textbf{Q: }Which of the boxes comes next?} \\ \tiny{\textbf{A: } E} \\ \tiny{\textbf{Reasoning Skill: } Inductive} \\ \tiny{\textbf{Capability: } Diagram} \end{tabular} & \raisebox{-.25\totalheight}{\includegraphics[width=2.65cm, height=2.65cm]{figures/vqav2.jpg}} & \begin{tabular}{c} \tiny{\textbf{Q: }Is the girl touching the ground?} \\ \tiny{\textbf{A: } No} \\ \tiny{\textbf{Reasoning Skill: } None} \\ \tiny{\textbf{Capability: } Recognition} \end{tabular} \\
    \\
    \raisebox{-.25\totalheight}{\includegraphics[width=2.65cm, height=2.65cm]{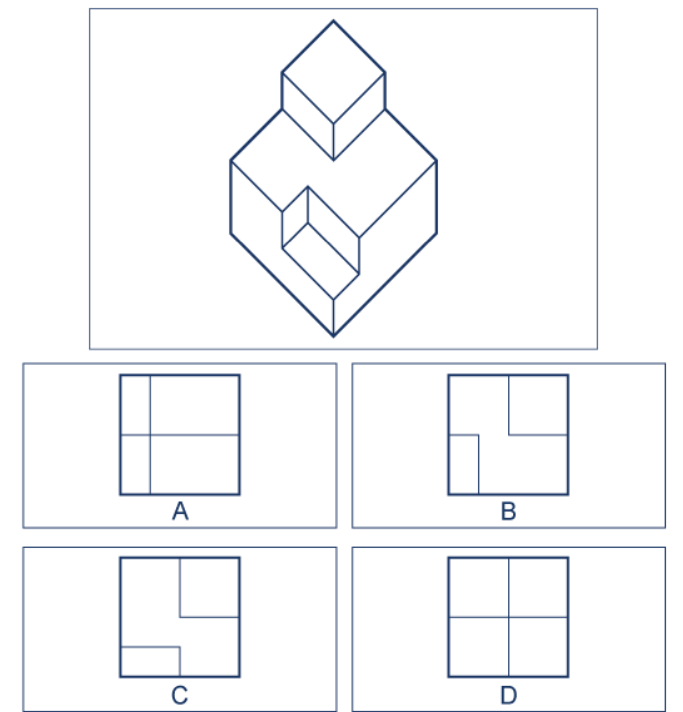}} & \begin{tabular}{c} \tiny{\textbf{Q: }Which of these are the top view?} \\ \tiny{\textbf{A: } B} \\ \tiny{\textbf{Reasoning Skill: }Spatial} \\ \tiny{\textbf{Capability: } 3D Shape} \end{tabular} & \raisebox{-.5\totalheight}{\includegraphics[width=2.65cm, height=2.65cm]{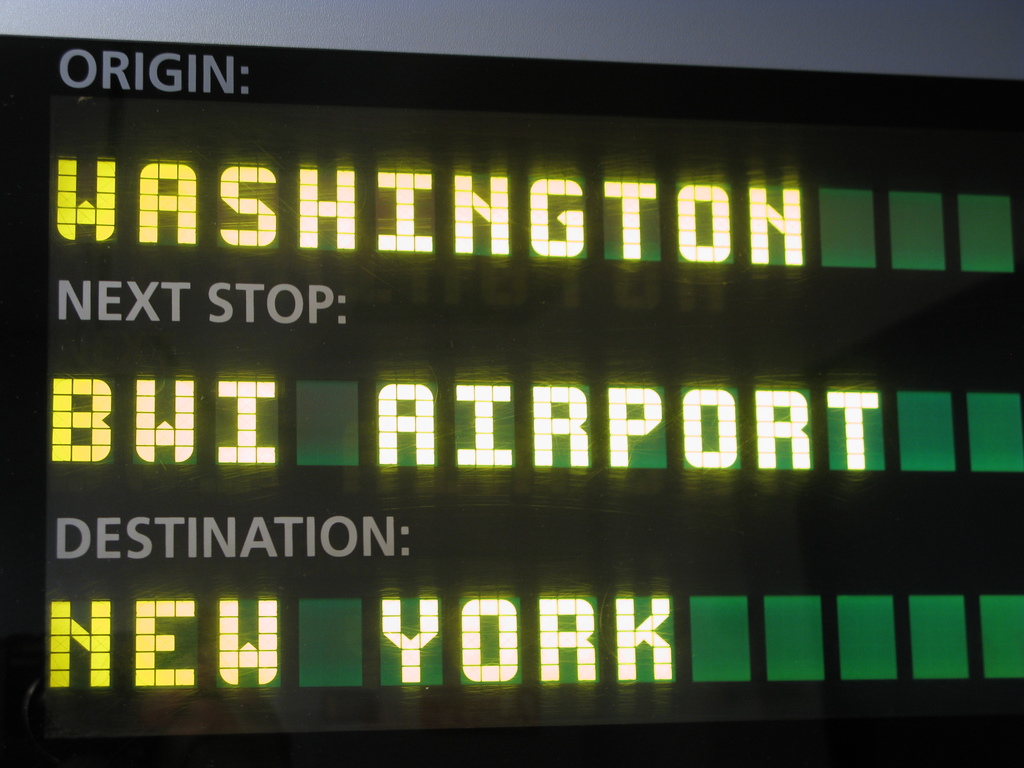}} & \begin{tabular}{c} \tiny{\textbf{Q: }What is the final destination?} \\ \tiny{\textbf{A: } New York} \\ \tiny{\textbf{Reasoning Skill: }None} \\ \tiny{\textbf{Capability: } OCR} \end{tabular} \\
    \\
    \raisebox{-.25\totalheight}{\includegraphics[width=2.65cm]{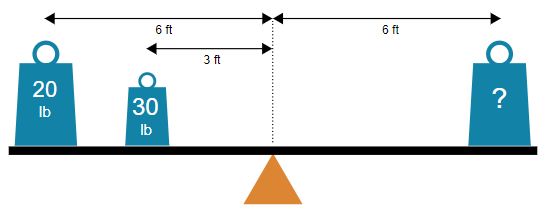}} & \begin{tabular}{c} \tiny{\textbf{Q: } What is the weight if balanced?} \\ \tiny{\textbf{A: } C: 35 lb} \\ \tiny{\textbf{Reasoning Skill: } Mechanical} \\ \tiny{\textbf{Capability: } Physics} \end{tabular} & \raisebox{-.5\totalheight}{\includegraphics[width=2.65cm, height=2.65cm]{figures/mmvet1.png}} & \begin{tabular}{c} \tiny{\textbf{Q: }What will girl on right write?} \\ \tiny{\textbf{A: } 14} \\ \tiny{\textbf{Reasoning Skill: }Numerical} \\ \tiny{\textbf{Capability: } OCR} \end{tabular} \\
    
\end{tabular}
\caption{Capabilities and reasoning skills of various existing benchmarks. Traditional benchmarks seldom assess reasoning skills, whereas \model emphasizes the fundamental capacities necessary for solving specific problems, going beyond simple recognition or math tasks.}
\label{fig:compare}
\end{center}
\end{figure*}

Recent MLLMs have demonstrated a range of impressive abilities, such as writing poems based on an image \citep{Fu}, engaging in mathematical reasoning \citep{alayrac2022flamingo}, and even aiding in medical diagnosis \citep{PMC-VQA}. To evaluate the performance of these models, various benchmarks have been proposed, as shown in Figure. \ref{fig:compare} targeting the performance on common tasks such as objects recognition \citep{VQA}, text understanding in images \citep{Text-VQA}, or mathematical problem solving \citep{yu2023mmvet}. However, as seen in Figure. \ref{fig:compare}, there is a notable shortage of benchmarks for MLLMs’ abilities in critical logical reasoning tasks that underlie most tasks. Perception and reasoning are two representative abilities of high-level intelligence that are used in unison during human problem-solving processes. 

Many current MLLM datasets have focused solely on perception tasks, which require fact retrieval where the MLLM identifies and retrieve relevant information from a scene. However, complex multimodal reasoning, such as interpreting graphs \citep{LineGraphs}, everyday reasoning, critical thinking, and problem-solving \citep{EarlyLogic, Bronkhorst} requires a combination of perception and logical reasoning. Proficiency in these reasoning skills is a reliable indicator of cognitive capabilities required for performing specialized or routine tasks across different domains. To our knowledge, MathVista \cite{lu2024mathvista} is the only benchmark that attempts to evaluate multimodal logical reasoning, but its scope is limited to mathematical-related reasoning. For a better understanding of how MLLMs perform on general reasoning tasks, there is a need for a comprehensive and general visual reasoning benchmark.

We argue that a universal comprehensive evaluation benchmark should have the following characteristics: (1) cover a wide range of logical reasoning tasks, including deductive, inductive, numeric, spatial, and mechanical reasoning; (2) present information in both graphical and Optical Character Recognition (OCR) formats to accommodate different types of data inputs; and (3) facilitate convenient quantitative analysis for rigorous assessment and comparison of model performance.

To this end, we present a comprehensive MLLM evaluation benchmark, named \modelNoSpace, which meets all these criteria:
\begin{itemize}
    \item \model covers 5 representative categories of logical reasoning tasks: inductive ($sample=\inductive$), deductive ($sample=\deductive$), numerical ($sample=\numerical$), spatial ($sample=\spatial$), and mechanical ($sample=\mech$).
    \item \model includes a variety of capabilities, ranging from diagrams ($sample=\diagram$), OCR, ($sample=\ocr$), patterns ($sample=\patternings$), graphs ($sample=\graphs$), tables ($sample=\tables$), 3D shapes ($samples=\dshapes$), puzzles ($samples=\puzzles$), sequences ($samples=\sequences$), and physics ($samples=\physics$).
    \item All images, instructions, solution, and reasoning are manually annotated and validated.
    \item With our instruction design ``please select from A, B, C, D, and E." and our LLM answer evaluator, we can assess different reasoning skills and capabilities and easily perform quantitative statistical analysis based on the natural language output of MLLMs. Additionally, We provide more in-depth human-written explanations for why each answer is correct, allowing for thorough open-ended evaluation. 
\end{itemize}

As shown in Figure. \ref{fig:compare}, \model covers a wide range of reasoning capabilities and evaluates them comprehensively. For instance, answering the question \textit{``Which of these images is the top view of the given object"} in Figure \ref{fig:compare}(b) requires not only recognizing the objects' orientation but also the ability to spatially reason over the object from a different perspective. Since these questions and diagrams are presented without context, they effectively probe the MLLM's underlying ability rather than relying on contextual cues from the surrounding real-life environment. 

Furthermore, we provide two evaluation strategies with our annotations: multiple-choice question (MCQ) evaluation and open-ended evaluation. Our annotation of MCQ choices along with our LLM evaluator allows quick evaluations of answers provided by MLLMs. Additionally, our annotation of the reasoning and thought process behind each MCQ enables open-ended evaluation, capturing the nuances of the MLLM responses and identifying which reasoning steps were correct or incorrect.

We comprehensively evaluate the performance of \MLLMcount representative open and closed source MLLMs on \datacount tasks across 5 main logical reasoning categories. \modelNoSpace's evaluation strategy allows users to see a detailed breakdown of an MLLM's performance on each reasoning skill and capability. This approach provides more insights than a single overall score, enabling users to better understand the specific skills in which a model excels or needs improvement.

\section{Related Works} \label{related}
\begin{table}[h]
\centering
\resizebox{\linewidth}{!}{
\begin{tabular}{lcccccccc}
\hline
& VQAv2~\cite{VQA, balanced_vqa_v2} & COCO ~\cite{COCO} & TextCaps~\cite{sidorov2020textcaps} & Contextual~\cite{contextual} & MM-vet~\cite{yu2023mmvet} & MathVista~\cite{lu2024mathvista} & VisIT-Bench~\cite{visit} & \textbf{\modelNoSpace} \\\hline
Number of Logical Reasoning Skills Tested & 0 & 0 & 1 & 1 & 1 & 2 & 1 & 5 \\
Number of Multimodal Capabilities Tested & 1 & 1 & 2 & 2 & 6 & 12 & 2 & 9\\
Dataset Size & 204,721 & 330,000 & 28,000 & 506 & 217 & 6,141 & 592 & 448 \\\hline
Scene and Object Recognition & \cmark & \cmark & \cmark & \cmark & \cmark & \cmark & \cmark & \cmark \\
Inductive Reasoning & \xmark & \xmark & \xmark & \xmark & \xmark & \cmark & \xmark & \cmark\\
Deductive Reasoning & \xmark & \xmark & \xmark & \xmark & \xmark & \xmark & \xmark & \cmark\\
Numerical Reasoning & \xmark & \xmark & \cmark & \cmark & \cmark & \cmark & \cmark & \cmark\\
Spatial Reasoning & \xmark & \xmark & \xmark & \xmark & \xmark & \xmark & \xmark & \cmark\\
Mechanical Reasoning & \xmark & \xmark & \xmark & \xmark & \xmark & \xmark & \xmark & \cmark\\ \hline
Answer Choice Explanations & \xmark & \xmark & \xmark & \xmark & \xmark & \cmark & \xmark & \cmark\\
Human Annotation & \cmark & \cmark & \cmark & \cmark & \cmark & \cmark & \cmark & \cmark\\
Human Evaluation & \xmark & \cmark & \cmark & \cmark & \xmark & \cmark & \cmark & \xmark\\
Auto/GPT-4 Evaluation & \cmark & \cmark & \cmark & \cmark & \cmark & \cmark & \cmark & \cmark\\
Open-ended Evaluation & \xmark & \cmark & \cmark & \cmark & \cmark & \cmark & \cmark & \cmark\\\hline
\end{tabular}
}
\caption{Comparision with related vision-language benchmarks.}
\label{relatedbench}
\end{table}

\noindent\textbf{Multimodal Language Models} 
The field of vision-language models~\citep{chen2015microsoft,goyal2017making,lu2019vilbert,chen2020uniter,li2020oscar,kim2021vilt,wang2022simvlm,wang2022git,yang2022unitab,gan2022visionlanguage} has made significant progress towards achieving a cohesive understanding and generation of both visual and linguistic information. This progress is largely driven by the remarkable generalization and quality capabilities of recent large language models (LLMs)~\citep{brown2020language,openai2024gpt4,chowdhery2022palm,touvron2023llama}. As a result, there has been a surge in the development of MLLMs that aim to integrate the diverse capabilities of vision and language for complex multimodal tasks.

Efforts to create these multimodal generalist systems include enhancing LLMs with multi-sensory processing abilities, as demonstrated by innovative projects like Frozen~\citep{tsimpoukelli2021multimodal}, Flamingo~\citep{alayrac2022flamingo}, PaLM-E~\citep{driess2023palme}, and GPT-4~\citep{openai2024gpt4}. Recent releases of open-source LLMs~\citep{zhang2022opt,touvron2023llama,peng2023instruction} have further propelled research in this field, leading to the development of OpenFlamingo~\citep{awadalla2023openflamingo}, LLaVA~\citep{liu2023visual}, MiniGPT-4~\citep{zhu2023minigpt4}, Otter~\citep{li2023otter}, InstructBLIP~\citep{dai2023instructblip}, among others~\citep{gong2023multimodalgpt,liu2023visual,ye2023mplugowl}. Additionally, multimodal agents~\citep{yang2023mmreact,shen2023hugginggpt,gao2023assistgpt} have been explored for their ability to link various vision tools with LLMs~\citep{brown2020language,openai2024gpt4}, aiming to enhance integrated vision-language capabilities

\noindent\textbf{Vision-Language Benchmarks}
Traditional vision-language benchmarks have focused on assessing specific capabilities, including visual recognition~\citep{goyal2017making}, generating image descriptions~\citep{chen2015microsoft, Agrawal_2019}, and other specialized functions such as understanding scene text~\citep{singh2019vqa,sidorov2020textcaps,yang2020tap}, commonsense reasoning~\citep{zellers2019recognition}, mathematical reasoning~\citep{lu2024mathvista}, instruction following ~\citep{visit}, and external knowledge incorporation~\citep{marino2019okvqa}. While some benchmarks incorporate reasoning ~\cite{contextual}, they are often presented in real-life contexts, which may reduce the task to mere recognition based on contextual cues. 

The emergence of general MLLMs has highlighted the need for updated vision-language benchmarks that encompass complex multimodal tasks requiring comprehensive vision-language skills. Our benchmark, \modelNoSpace, aligns closely with recent evaluation studies like MM-Vet and MMBench~\citep{yu2023mmvet,liu2023mmbench}, which aim to provide thorough evaluations of MLLMs through well-designed evaluation samples. A key distinction of \model lies in its focus on integrated vision-language capabilities, offering deeper insights beyond mere model rankings.

\noindent\textbf{LLM-Based Evaluation.}
\model adopts an open-ended LLM-based evaluation approach, which facilitates the generation and assessment of diverse answer styles and question types beyond the limitations of binary or multiple-choice responses. This innovative method leverages the capabilities of large language models (LLMs) for comprehensive model evaluation, a technique that has been effectively applied in natural language processing (NLP) tasks~\citep{chiang2023large,liu2023geval,fu2023gptscore, jin2024mm}. Our findings indicate that this LLM-based evaluation framework is not only versatile but also robust, enabling a unified and flexible assessment across various modalities. By accommodating a wide range of answer styles and question types, this approach enhances evaluation depth and breadth, which contributes to a more thorough understanding of model performance.

\section{Data annotation and organization} \label{method}
\begin{figure}
     \centering
     \begin{subfigure}[b]{0.35\textwidth}
         \centering
         \includegraphics[height=5cm]{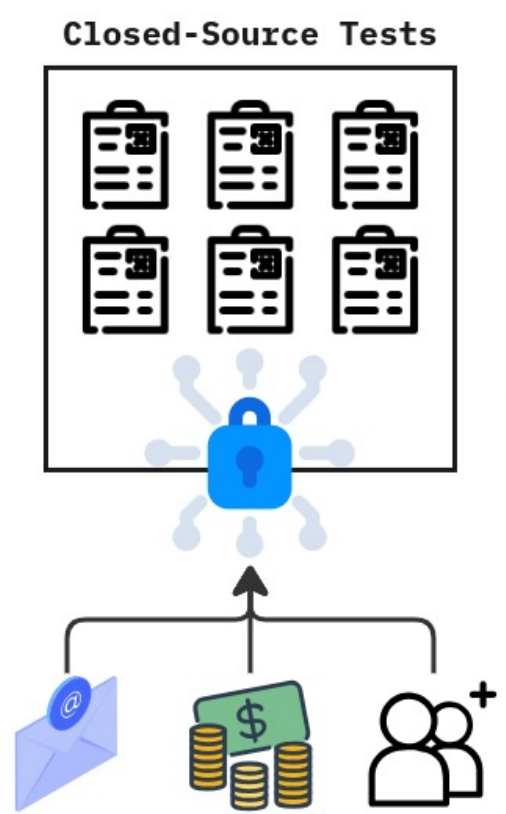}
         \caption{}
         \label{fig:datasource1}
     \end{subfigure}
    \hspace{0.05\textwidth} 
     \begin{subfigure}[b]{0.35\textwidth}
         \centering
         \includegraphics[height=5cm]{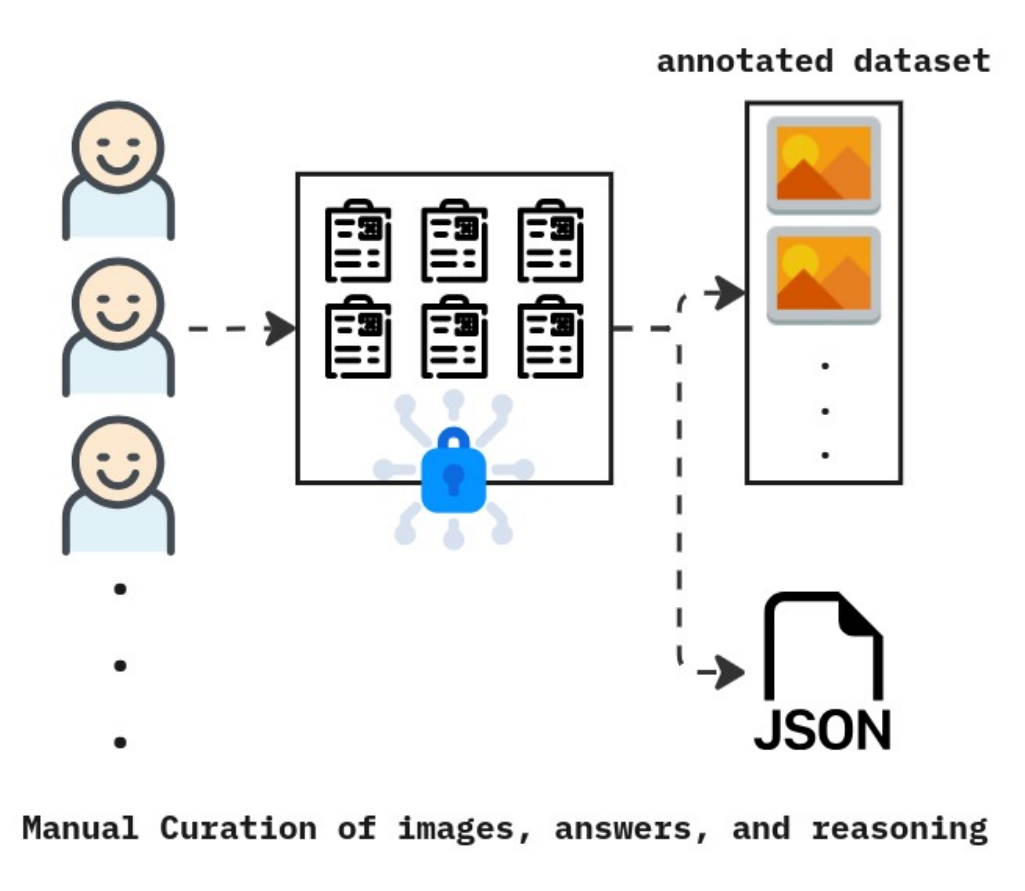}
         \caption{}
         \label{fig:datasource2}
     \end{subfigure}
     \caption{a) Data collected for \model were gathered from closed sources to avoid data leakage. b) Manual annotators used the gathered tests, gathered the correct answers, and came up with reasonings on why the selected answers were correct. All these annotations were then stored in JSON format.}
     \label{fig:datasource}
\end{figure}

\subsection{Data Sources}
To ensure the integrity and quality of \modelNoSpace’s evaluations, we have implemented a stringent data collection and curation process specifically designed to prevent data leakage detailed in Figure. \ref{fig:datasource}. Our approach involves sourcing and annotating our samples from proprietary sources that require licenses, registration, payment, or a combination of these barriers to access. This methodology is critical to minimizing the risk that our benchmark data has been previously seen or utilized in the training of other multi-modal models. We prioritized sourcing data from closed sources to further reduce the potential of data leakage.
\begin{itemize}
    \item \textbf{Licensed Access:} We obtain data from sources that require formal licensing, ensuring the data is used solely for research purposes and not freely available for general use or scraping on the internet.
    \item \textbf{Registration Requirements:} Some of our data sources mandate user registration and account verification, adding an additional layer of access control to ensure that the data remain restricted and not easily accessible. 
    \item \textbf{Paid Content:} We utilize paid sources where content is accessible only through purchase or subscription, further restricting the data from being freely available on the internet. 
\end{itemize}

Additionally, we obtained permission from the creators of IQ tests and other evaluation materials included in our dataset. This permission specifically allows the use of their content for research purposes, ensuring the data's legitimacy and accuracy.

\subsection{Annotation and Data Collection}
\model consists of images designed to assess the underlying reasoning capacities of MLLMs. Using real-life scenes as explicit tests of logical reasoning can be challenging, as they often contain context clues that AI agent can use to deduce answers without directly reasoning through the scene. Therefore, \model presents multiple-choice questions across \capcount explicit capabilities that specify the type of reasoning required, without the additional context of real-life scenes typically found in intelligence and reasoning tests. The dataset is manually collected and annotated from various licensed intelligence test sources. Over a period of 3 months, 5 annotators extracted images, correct answers, and explanations when available. The explanations detailing the reasoning behind answer choices were extensively annotated and cross-validated among annotators, ensuring data integrity through multiple rounds of quality checks. The data is structured in JSON format to facilitate easy retrieval and processing in our evaluation pipeline. For our evaluation, we focused on summarizing five reasoning skills spanning two multimodal capabilities. For detailed examples of these reasoning skills and capabilities, please refer to Appendix. \ref{app:log} and Appendix. \ref{app:cap}.

\begin{figure*}[tbp]
\centering  \includegraphics[width=0.9\linewidth]{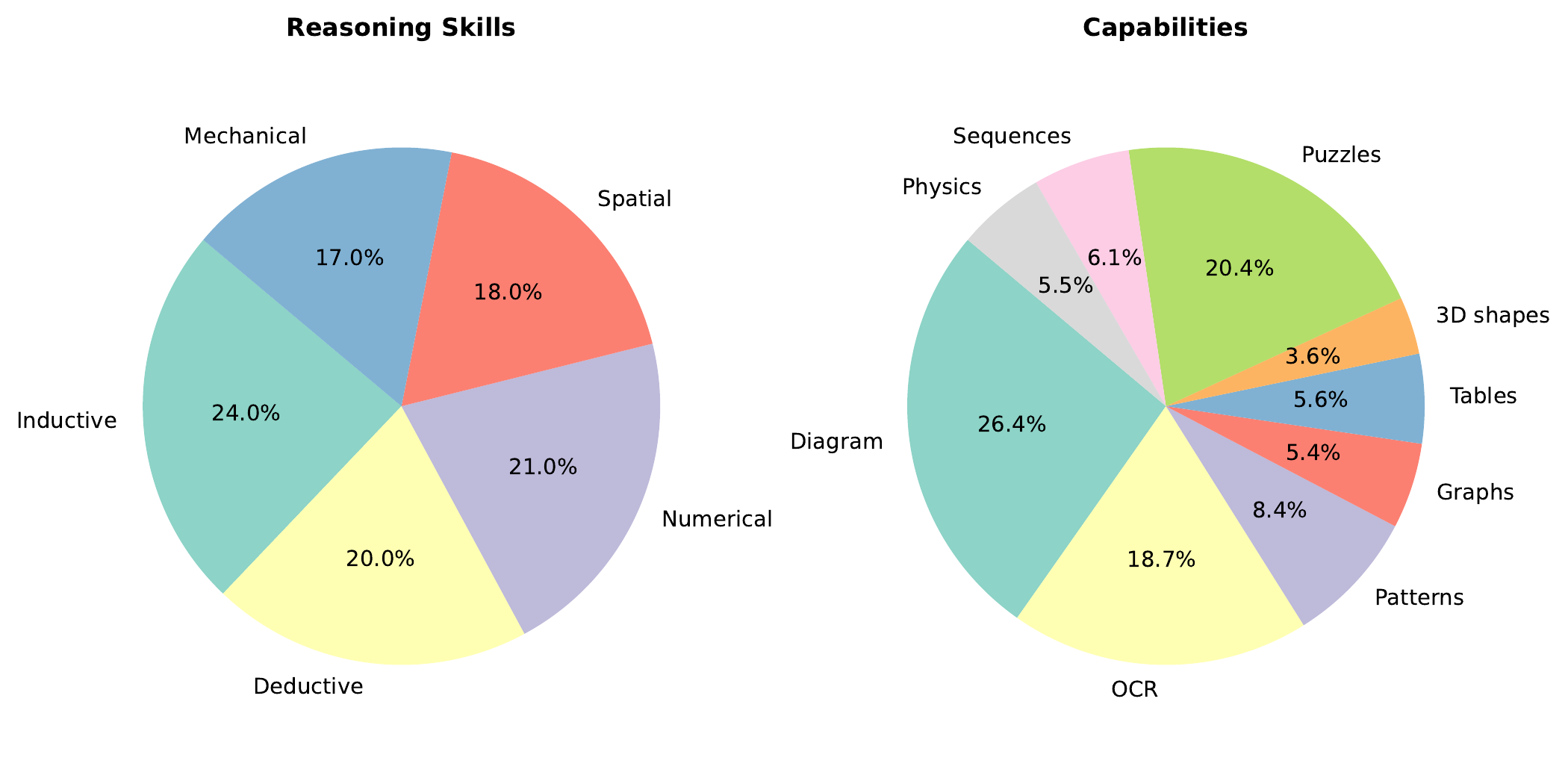}
  \caption{Proportion of reasoning skills and capabilities. On the left is the proportion of questions belonging to each reasoning skill. These proportions add up to $100\%$ as each skill is independent of another. On the right is the proportion of questions belonging to each multi-modal capability. These do not add up to $100\%$ due to the use of mixed capabilities.}
  \label{fig:bar_graph}
\end{figure*}

\subsubsection{Capabilities}
We distinguish multimodal capabilities from reasoning skills, considering these capabilities fundamental to understanding a multimodal scene and extracting information. Capabilities refer to the modalities through which logical reasoning questions are delivered. To ensure comprehensive coverage in \modelNoSpace, we have defined a diverse array of \capcount capabilities for evaluation. This diversity guarantees that \model thoroughly assess various logical situations that an MLLM may encounter in everyday reasoning.
Figure \ref{fig:bar_graph} demonstrates how \model contains a balanced mix of capabilities, including samples that utilize multiple capabilities to solve a problem.

\begin{itemize}
    \item \textbf{Diagrams:} Simple flow diagrams and logical diagrams (e.g., Markov diagrams).
    \item \textbf{OCR:} Text embedded within an image (e.g., ``gas station'' in an image of a gas station).
    \item \textbf{Patterns:} Repeated sequences such as a series of diagrams, numbers, shapes, and objects (e.g., identifying patterns in how a box moves through repeated images of boxes).
    \item \textbf{Graphs:} Mathematical graphs with axes (e.g., graphs of \(y=2x\) and \(y = x^2\)).
    \item \textbf{Tables:} Data tables (e.g., pie charts and T-tables).
    \item \textbf{3D Shapes:} The ability to understand and differentiate 3D objects from 2D ones (e.g., recognizing a 3D shape in different rotations).
    \item \textbf{Puzzles:} Puzzles with logical implications embedded within the shapes (e.g., chess puzzles).
    \item \textbf{Sequences:} Sequences of related items or objects (e.g., predicting the next item in a sequence).
    \item \textbf{Physics:} Situations involving physics (e.g., diagrams of projectile motion).
\end{itemize}

\subsubsection{Reasoning Skills}
The reasoning skills of interest for this benchmark are based on common critical thinking and problem-solving skills used by humans in various contexts. For our evaluation, we summarize these into the following five skills. For our evaluation, we summarize these to include the following 5 skills. As seen in Figure \ref{fig:bar_graph}, \model encompasses a wide range of all these reasoning skills:

\begin{itemize}
    \item \textbf{Inductive Reasoning}: 
    The ability to infer the next entry in a pattern given a set of observations. This involves making generalizations based on specific observations to form an educated guess.  It moves from many specific observations to a generalization. For example, observing that John gets a stomach ache when he eats dairy products leads to the inductive conclusion that he is likely lactose intolerant. 
    \item \textbf{Deductive Reasoning}:
    The ability to conclude a specific case from a general principle or pattern. This involves moving from the general to the specific. For example, from the statement “all men are mortal,” one can deduce that “John is mortal” because John is a man.
     \item \textbf{Numerical Reasoning}:
    The ability to read arithmetic problems in an image and solve the math equations. For example, given the equation “10 + 10 = ?,” the answer would be “20.”
    \item \textbf{Spatial Reasoning}:
    The ability to understand the spatial relationships between objects and patterns and reason with those relationships. For example, seeing an unfolded box and understanding what the box would look like when folded.
    \item \textbf{Mechanical Reasoning}:
    The ability to recognize a physical system and solve equations based on that system or answer questions about it. For example, seeing a set of three gears and understanding which gears will turn clockwise and which will turn counterclockwise.
\end{itemize}

\subsection{LLM-based Multiple Choice Answer Extractor}
\begin{figure}[tbp]
         \centering
         \includegraphics[height=5.5cm]{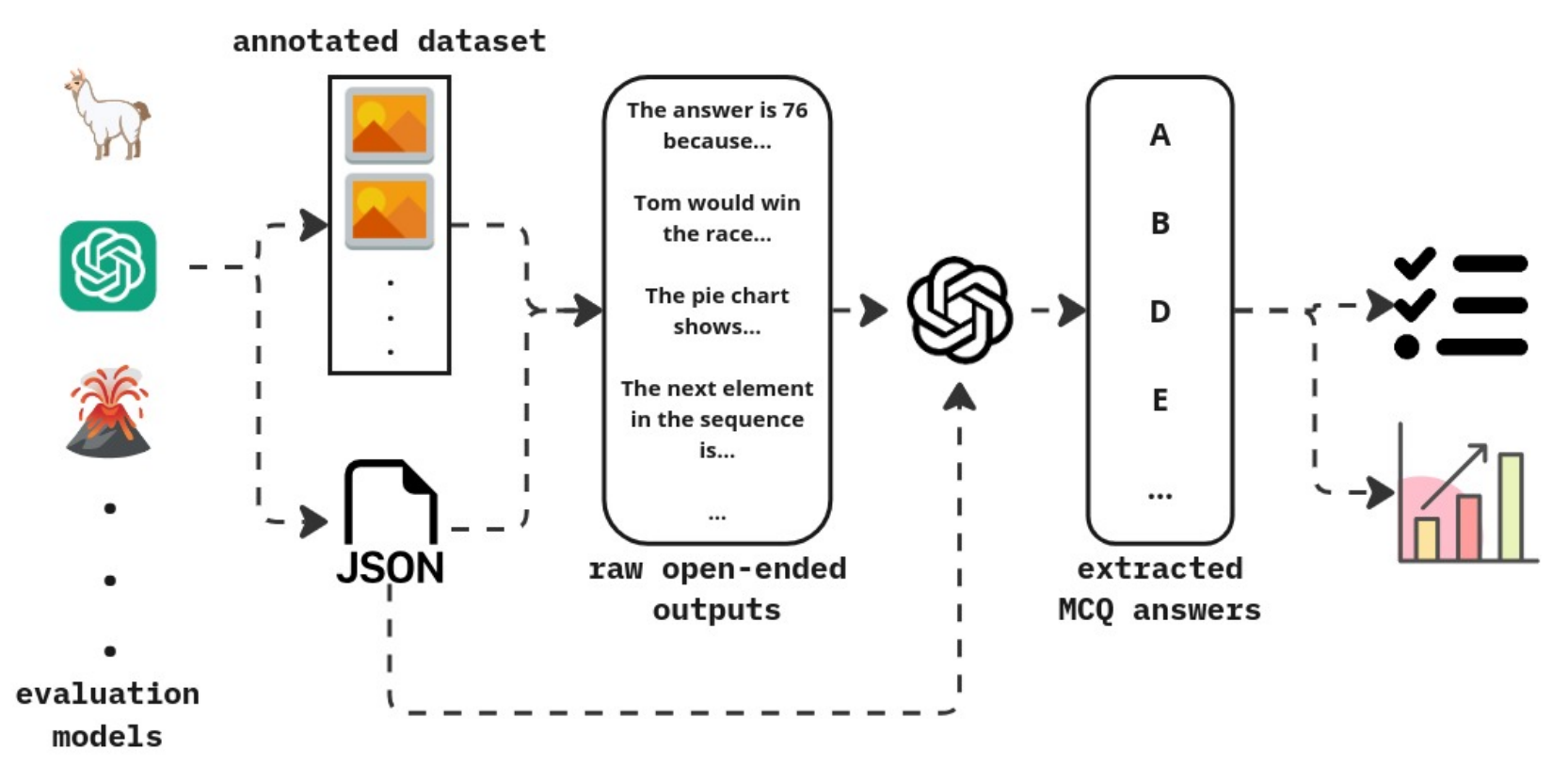}
     \caption{Pipeline of evaluating open-ended LMM outputs using MCQ answer choice extraction.}
     \label{fig:mcqextract}
\end{figure}
LLMs generate non-deterministic and open-ended responses \citep{Lee_2022, ouyang2023llm}, making direct evaluation challenging. To address this, we use an LLM evaluator to compare these open-ended responses to our annotations as detailed in \ref{fig:mcqextract}. This evaluator can assess both MCQ answer choices and the MLLM's reasoning behind those selections, as both elements are included in our annotations. This step is achieved by feeding various contexts such as the question, and the available choices, along with the LLM-generated answers to an extraction LLM (GPT, LLaMA, etc.). Based on the provided rich context, the LLM can generate the selected letter answer choice. The final output is also repeatedly validated and if the validation fails, the extraction repeats with the provided feedback to obtain correct results.

\section{Evaluation Setup} \label{setup}
\begin{table*}[htbp]
\label{modelsummary}
\begin{center}
\resizebox{0.8\textwidth}{!}{%
\begin{tabular}{llll}
\toprule
\bf Model  & \bf Size &\bf Language Model  &\bf Vision Model \\
\midrule
LLaVA-Vicuna-7B & 7B & Vicuna-7B & CLIP ViT-L/14 \\
LLaVA-Vicuna-13B & 13B & Vicuna-13B & CLIP ViT-L/336px \\
\midrule
LLaVA-NeXT-Mistral-7B & 7B & Mistral-7B & CLIP ViT-L/14 \\
LLaVA-NeXT-Vicuna-7B & 7B & Vicuna-7B & CLIP ViT-L/14 \\
LLaVA-NeXT-Vicuna-13B & 13B & Vicuna-13B & CLIP ViT-L/336px \\
LLaVA-NeXT-Nous-Hermes-Yi-34B & 34B & Nous Hermes 2-Yi-34B & CLIP ViT-L/336px \\
\midrule
MiniGPT-4-7B & 7B & Vicuna-7B & BLIP-2 Q-Former \\
MiniGPT-4-13B & 13B & Vicuna-13B & BLIP-2 Q-Former \\
\midrule
Otter-9B & 9B & MPT-7B & CLIP ViT-L/14 \\
\midrule
GPT-4 Vision & N/A\tablefootnote{N/A: Not disclosed} & N/A & N/A \\
\midrule
BLIP-2 & 2.7B & OPT-2.7B & EVA-ViT-G \\
\midrule
Pix2Struct & 1.3B & ViT & ViT \\
\midrule
InstructBLIP-Vicuna-7B & 7B &Vicuna-7B & BLIP-2 Q-Former \\
InstructBLIP-Vicuna-13B & 13B &Vicuna-13B & BLIP-2 Q-Former \\
InstructBLIP-FLAN-T5-xl & 3B &FLAN-T5 XL& BLIP-2 Q-Former \\
InstructBLIP-FLAN-T5-xxl & 11B &FLAN-T5 XXL& BLIP-2 Q-Former \\
\bottomrule
\end{tabular}
}
\end{center}
\caption{Summary of the MLLMs used for evaluations in this study.} 
\label{table:MLLMsummary}
\end{table*}

To evaluate the performance of MLLMs on \modelNoSpace, we selected a range of representative models detailed in Table. \ref{table:MLLMsummary}. Specifically, we chose\MLLMcount models for evaluation, including LLaVA \citep{liu2023llava, liu2024llavanext}, MiniGPT4 \citep{zhu2023minigpt4}, Otter \citep{li2023otter}, GPT-4 Vision \citep{openai2024gpt4}, BLIP-2 \citep{li2023blip2}, and InstructBLIP \citep{dai2023instructblip} We also included pix2struct \citep{lee2023pix2struct} which has been fine-tuned to understand chart and diagram data.


Each model generated outputs using the \model dataset. Our LLM-based multiple-choice extractor was then employed to isolate the multiple-choice selections from the MLLMs' outputs (which often appear as full-sentence responses rather than single letters) and compare them to the ground truth answers. The overall logical reasoning score is calculated as follows:

\begin{equation}
S = \frac{\sum_{n=1}^{N} s_i}{N}  * 100\%
\end{equation}

Here, $S$ represents the overall score, $s_i$ indicate whether a sample $i$ is evaluated as correct or not (regardless of category), and $N$ is the total number of samples. The score for each reasoning skill subcategory is calculated as:

\begin{equation}
S_{LR} = \frac{\sum_{n=1}^{N_{LR}} s_i}{N_{LR}}  * 100\%
\end{equation}

where $S_{LR}$ represents the score for a specific reasoning skill category, $N_{LR}$ is the total number of samples in that reasoning skill category, and $s_i$ indicate whether a sample $i$ from that category was evaluated as correct. Similarly, the score for each multi-modal capability is calculated as:

\begin{equation}
S_{c} = \frac{\sum_{n=1}^{N_{c}} s_i}{N_{c}}  * 100\%
\end{equation}

where $S_{c}$ represents the score for a specific capability, $N_{c}$ is the total number of samples in that capability, and $s_i$ indicates whether a sample $i$ in the capability category is evaluated correctly.

\section{\model Benchmarking and Performance Interpretation} \label{results}

\subsection{Logical Reasoning Skills}
We present the performance results of various multimodal LLMs on \modelNoSpace. Table \ref{tab:reason_skill} outlines the outcome for these models across five logical reasoning categories. We analyzed models of different architectures and sizes, benchmarking them against a random baseline that assumes an average of five choices per question in the \model dataset. Our findings indicate that many models perform below expectations, often yielding results that are worse than random guessing. This outcome is somewhat anticipated, given that most training data for multimodal LLMs and LLMs are derived from classical computer vision datasets such as COCO, which focus on recognition tasks rather than complex reasoning.

Traditional benchmarks typically emphasize recognition tasks, resulting in a lack of emphasis on reasoning tasks during both training and evaluation phases. This is evident from the observation that while many models excel on recognition-based benchmarks like COCO, TextVQA, and MM-vet, they often struggle to outperform a random baseline on logical reasoning tasks.

\begin{table*}[htbp]
\scriptsize
  \centering
  \resizebox{0.9\textwidth}{!}{%
    \begin{tabular}{lcccccc}
    \toprule
    \textbf{Model} & \textbf{Inductive} & \textbf{Deductive} & \textbf{Numerical} & \textbf{Spatial} & \textbf{Mechanical} \\
    \midrule
    \textbf{LLAVA7B} & \cellcolor[rgb]{ .435,  .765,  .58} 29.91\% & \cellcolor[rgb]{ .89,  .914,  .757} 29.03\% & \cellcolor[rgb]{ .718,  .859,  .69} 26.32\% & \cellcolor[rgb]{ .584,  .812,  .635} 25.32\% & \cellcolor[rgb]{ .537,  .8,  .62} 36.49\% \\
    \textbf{LLAVA13B} & \cellcolor[rgb]{ .98,  .945,  .792} 18.69\% & \cellcolor[rgb]{ .843,  .898,  .741} 31.18\% & \cellcolor[rgb]{ 1,  .957,  .831} 20.00\% & \cellcolor[rgb]{ .341,  .733,  .541} 27.85\% & \cellcolor[rgb]{ .976,  .941,  .792} 24.32\% \\
    \textbf{otter9B} & \cellcolor[rgb]{ .341,  .733,  .541} 31.78\% & \cellcolor[rgb]{ .98,  .945,  .792} 24.73\% & \cellcolor[rgb]{ 1,  .961,  .843} 18.95\% & \cellcolor[rgb]{ 1,  .957,  .824} 18.99\% & \cellcolor[rgb]{ 1,  .969,  .871} 21.62\% \\
    \textbf{GPT4} & \cellcolor[rgb]{ .753,  .871,  .702} 23.36\% & \cellcolor[rgb]{ .341,  .733,  .541} 54.84\% & \cellcolor[rgb]{ .91,  .922,  .765} 24.21\% & \cellcolor[rgb]{ .941,  .929,  .776} 21.52\% & \cellcolor[rgb]{ .341,  .733,  .541} 41.89\% \\
    \textbf{BLIP2} & \cellcolor[rgb]{ 1,  .953,  .808} 17.76\% & \cellcolor[rgb]{ 1,  .949,  .8} 23.66\% & \cellcolor[rgb]{ 1,  .949,  .8} 23.16\% & \cellcolor[rgb]{ .702,  .855,  .686} 24.05\% & \cellcolor[rgb]{ 1,  .992,  .957} 18.92\% \\
    \textbf{LLAVANEXT-7B-mistral} & \cellcolor[rgb]{ 1,  .957,  .82} 16.82\% & \cellcolor[rgb]{ .776,  .878,  .714} 34.41\% & \cellcolor[rgb]{ 1,  .953,  .804} 23.16\% & \cellcolor[rgb]{ .941,  .933,  .78} 21.52\% & \cellcolor[rgb]{ 1,  .957,  .824} 22.97\% \\
    \textbf{miniGPTvicuna7B} & \cellcolor[rgb]{ 1,  .98,  .91} 10.28\% & \cellcolor[rgb]{ 1,  .992,  .965} 9.68\% & \cellcolor[rgb]{ 1,  .988,  .953} 7.37\% & \cellcolor[rgb]{ 1,  1,  .988} 3.80\% & \cellcolor[rgb]{ .878,  .91,  .753} 27.03\% \\
    \textbf{miniGPTvicuna13B} & \cellcolor[rgb]{ 1,  .969,  .875} 13.08\% & \cellcolor[rgb]{ 1,  .953,  .804} 23.66\% & \cellcolor[rgb]{ 1,  .98,  .922} 10.53\% & \cellcolor[rgb]{ 1,  .98,  .918} 10.13\% & 17.57\% \\
    \textbf{pix2struct} & \cellcolor[rgb]{ 1,  .973,  .886} 12.15\% & 6.45\% & 2.11\% & \cellcolor[rgb]{ 1,  .988,  .945} 7.59\% & 17.57\% \\
    \textbf{instructBLIP-vicuna-7B} & \cellcolor[rgb]{ 1,  1,  .988} 4.67\% & \cellcolor[rgb]{ 1,  .957,  .827} 21.51\% & \cellcolor[rgb]{ .91,  .922,  .765} 24.21\% & 2.53\% & \cellcolor[rgb]{ 1,  .957,  .824} 22.97\% \\
    \textbf{instructBLIP-vicuna-13B} & 3.74\% & \cellcolor[rgb]{ 1,  .988,  .953} 10.75\% & \cellcolor[rgb]{ 1,  .961,  .843} 18.95\% & \cellcolor[rgb]{ 1,  .996,  .973} 5.06\% & 17.57\% \\
    \textbf{instructBLIP-flan-t5-xl} & \cellcolor[rgb]{ .753,  .871,  .702} 23.36\% & \cellcolor[rgb]{ 1,  .953,  .816} 22.58\% & \cellcolor[rgb]{ 1,  .953,  .812} 22.11\% & \cellcolor[rgb]{ 1,  .988,  .945} 7.59\% & \cellcolor[rgb]{ .635,  .831,  .659} 33.78\% \\
    \textbf{instructBLIP-flan-t5-xxl} & \cellcolor[rgb]{ 1,  .953,  .808} 17.76\% & \cellcolor[rgb]{ .867,  .906,  .749} 30.11\% & \cellcolor[rgb]{ .91,  .922,  .765} 24.21\% & \cellcolor[rgb]{ 1,  .953,  .808} 20.25\% & \cellcolor[rgb]{ 1,  .957,  .824} 22.97\% \\
    \textbf{LLAVANEXT-7B-vicuna} & \cellcolor[rgb]{ .616,  .824,  .651} 26.17\% & \cellcolor[rgb]{ 1,  .957,  .827} 21.51\% & \cellcolor[rgb]{ .816,  .89,  .729} 25.26\% & \cellcolor[rgb]{ .345,  .737,  .545} 27.85\% & \cellcolor[rgb]{ .78,  .878,  .714} 29.73\% \\
    \textbf{LLAVANEXT-13B-vicuna} & \cellcolor[rgb]{ .796,  .882,  .722} 22.43\% & \cellcolor[rgb]{ 1,  .953,  .816} 22.58\% & \cellcolor[rgb]{ .722,  .859,  .69} 26.32\% & \cellcolor[rgb]{ .463,  .776,  .592} 26.58\% & \cellcolor[rgb]{ .929,  .925,  .773} 25.68\% \\
    \textbf{LLAVANEXT-34B-NH} & \cellcolor[rgb]{ .89,  .914,  .757} 20.56\% & \cellcolor[rgb]{ .388,  .749,  .561} 52.69\% & \cellcolor[rgb]{ .341,  .733,  .541} 30.53\% & \cellcolor[rgb]{ .702,  .855,  .686} 24.05\% & \cellcolor[rgb]{ .392,  .753,  .561} 40.54\% \\
    \bottomrule
    \end{tabular}%
    }
    \caption{\model evaluation results for various multimodal LLMs on each logical reasoning skill are presented as $\%$, with the highest possible accuracy being $100\%$. The highest-scoring models are highlighted in \textcolor{green}{\textbf{green}} and the lower-scoring models are highlighted in \textcolor{olive}{\textbf{yellow}}.}
    \label{tab:reason_skill}%
\end{table*}%

Upon closer examination, we find that models perform best on deductive, numerical, and mechanical reasoning tasks. These types of reasoning are more prevalent in real-life scenarios, which makes models more adept at handling them. For example, deductive reasoning can be applied in predicting a character's actions based on a scene, while numerical reasoning is crucial in solving arithmetic visual tasks. Mechanical reasoning involves understanding physical principles and interactions.

In contrast, induction and spatial reasoning are less frequently encountered in standard training data, potentially explaining the lower performance of models in these areas. These insights underscore the necessity for enhanced training and evaluation methodologies that prioritize reasoning tasks to bolster the logical reasoning capabilities of multimodal LLMs.

\subsection{Visual Capabilities}

In Table \ref{tab:capability}, we present the results of multimodal LLMs on logical reasoning tasks across diagrammatic and OCR mediums. Generally, we observe that OCR tasks tend to perform better than diagrammatic tasks. This difference stems from the nature of traditional computer vision tasks, which often prioritize recognizing prominent objects (``landmarks'') in a scene, such as distinct cars, planes, people, or balls. Diagrams, in contrast, lack such prominent features and mainly consist of lines and shapes, making it challenging for models to extract intricate relationships between objects.

In OCR tasks, once the text is accurately extracted from the image, the remainder of the reasoning task relies on the underlying LLM's ability to process and interpret the content. This process typically bypasses the complexities of multimodal reasoning, leading to better performance on OCR tasks compared to diagrammatic tasks. These findings highlight the necessity for enhanced evaluation methodologies tailored to diagrammatic reasoning in multimodal LLMs, as current approaches may overlook critical details inherent in these types of tasks.

\begin{table*}[htbp]
  \centering
  \resizebox{1\textwidth}{!}{%
    \begin{tabular}{lccccccccc}
    \toprule
    \textbf{Model} & \textbf{Diagram} & \textbf{OCR} & \textbf{Patterns} & \textbf{Graphs} & \textbf{Tables} & \textbf{3D Shapes} & \textbf{Puzzles} & \textbf{Sequences} & \textbf{Physics} \\
    \midrule
    \textbf{LLAVA7B} & \cellcolor[rgb]{ .341,  .733,  .541} 29.70\% & \cellcolor[rgb]{ .78,  .878,  .714} 28.21\% & \cellcolor[rgb]{ .345,  .737,  .545} 30.47\% & \cellcolor[rgb]{ .584,  .816,  .639} 25.37\% & \cellcolor[rgb]{ .816,  .89,  .729} 25.71\% & \cellcolor[rgb]{ .757,  .871,  .706} 22.22\% & \cellcolor[rgb]{ .545,  .8,  .624} 28.52\% & \cellcolor[rgb]{ .541,  .8,  .62} 25.00\% & \cellcolor[rgb]{ .471,  .776,  .592} 43.48\% \\
    \textbf{LLAVA13B} & \cellcolor[rgb]{ .937,  .929,  .776} 21.52\% & \cellcolor[rgb]{ .992,  .949,  .8} 22.65\% & \cellcolor[rgb]{ 1,  .965,  .859} 16.19\% & \cellcolor[rgb]{ 1,  .961,  .839} 16.42\% & \cellcolor[rgb]{ 1,  .961,  .831} 20.00\% & \cellcolor[rgb]{ .424,  .761,  .576} 31.11\% & \cellcolor[rgb]{ .718,  .859,  .69} 26.17\% & \cellcolor[rgb]{ 1,  .949,  .8} 15.79\% & \cellcolor[rgb]{ .98,  .945,  .792} 26.09\% \\
    \textbf{otter9B} & \cellcolor[rgb]{ .78,  .878,  .714} 23.64\% & \cellcolor[rgb]{ 1,  .957,  .831} 20.51\% & \cellcolor[rgb]{ .341,  .733,  .541} 30.48\% & \cellcolor[rgb]{ 1,  .965,  .855} 14.93\% & \cellcolor[rgb]{ 1,  .949,  .8} 22.86\% & \cellcolor[rgb]{ 1,  .965,  .851} 13.33\% & \cellcolor[rgb]{ .718,  .859,  .69} 26.17\% & \cellcolor[rgb]{ .475,  .776,  .596} 26.32\% & \cellcolor[rgb]{ 1,  .953,  .816} 24.64\% \\
    \textbf{GPT4} & \cellcolor[rgb]{ .608,  .824,  .647} 26.06\% & \cellcolor[rgb]{ .341,  .733,  .541} 39.74\% & \cellcolor[rgb]{ .973,  .941,  .788} 20.95\% & \cellcolor[rgb]{ .941,  .929,  .776} 20.90\% & \cellcolor[rgb]{ 1,  .949,  .8} 22.86\% & \cellcolor[rgb]{ .424,  .761,  .576} 31.11\% & \cellcolor[rgb]{ .341,  .733,  .541} 31.25\% & \cellcolor[rgb]{ .341,  .733,  .541} 28.95\% & \cellcolor[rgb]{ .341,  .733,  .541} 47.83\% \\
    \textbf{BLIP2} & \cellcolor[rgb]{ 1,  .953,  .808} 20.30\% & \cellcolor[rgb]{ 1,  .953,  .812} 21.79\% & \cellcolor[rgb]{ 1,  .953,  .808} 20.00\% & \cellcolor[rgb]{ 1,  .957,  .824} 17.91\% & \cellcolor[rgb]{ .906,  .922,  .765} 24.29\% & \cellcolor[rgb]{ .918,  .925,  .769} 17.78\% & \cellcolor[rgb]{ 1,  .949,  .8} 22.27\% & \cellcolor[rgb]{ 1,  .949,  .8} 15.79\% & \cellcolor[rgb]{ 1,  .976,  .894} 20.29\% \\
    \textbf{LLAVANEXT-7B-mistral} & \cellcolor[rgb]{ 1,  .953,  .808} 20.30\% & \cellcolor[rgb]{ .831,  .894,  .733} 26.92\% & \cellcolor[rgb]{ .91,  .922,  .765} 21.90\% & \cellcolor[rgb]{ .702,  .855,  .686} 23.88\% & \cellcolor[rgb]{ 1,  .949,  .8} 22.86\% & \cellcolor[rgb]{ 1,  .965,  .851} 13.33\% & \cellcolor[rgb]{ 1,  .949,  .8} 22.27\% & \cellcolor[rgb]{ .608,  .824,  .647} 23.68\% & \cellcolor[rgb]{ .855,  .902,  .745} 30.43\% \\
    \textbf{miniGPTvicuna7B} & \cellcolor[rgb]{ 1,  .996,  .976} 10.91\% & \cellcolor[rgb]{ 1,  .992,  .961} 11.54\% & \cellcolor[rgb]{ 1,  .98,  .91} 12.38\% & \cellcolor[rgb]{ 1,  .984,  .929} 7.46\% & \cellcolor[rgb]{ 1,  .992,  .957} 8.57\% & \cellcolor[rgb]{ 1,  .976,  .902} 11.11\% & \cellcolor[rgb]{ 1,  .992,  .957} 9.77\% & \cellcolor[rgb]{ 1,  .976,  .902} 7.89\% & \cellcolor[rgb]{ 1,  .961,  .843} 23.19\% \\
    \textbf{miniGPTvicuna13B} & \cellcolor[rgb]{ 1,  .988,  .941} 12.73\% & \cellcolor[rgb]{ 1,  .969,  .875} 17.52\% & \cellcolor[rgb]{ 1,  .98,  .91} 12.38\% & \cellcolor[rgb]{ 1,  .976,  .898} 10.45\% & \cellcolor[rgb]{ 1,  .984,  .925} 11.43\% & \cellcolor[rgb]{ 1,  .976,  .902} 11.11\% & \cellcolor[rgb]{ 1,  .976,  .894} 14.84\% & \cellcolor[rgb]{ 1,  .98,  .918} 6.58\% & \cellcolor[rgb]{ 1,  .976,  .894} 20.29\% \\
    \textbf{pix2struct} & 9.39\% & 8.55\% & \cellcolor[rgb]{ 1,  .984,  .937} 10.48\% & 0.00\% & 4.29\% & \cellcolor[rgb]{ 1,  .976,  .902} 11.11\% & \cellcolor[rgb]{ 1,  .988,  .949} 10.55\% & \cellcolor[rgb]{ 1,  .965,  .851} 11.84\% & 14.49\% \\
    \textbf{instructBLIP-vicuna-7B} & \cellcolor[rgb]{ 1,  .992,  .957} 11.82\% & \cellcolor[rgb]{ 1,  .957,  .816} 21.37\% & \cellcolor[rgb]{ 1,  .996,  .976} 7.62\% & \cellcolor[rgb]{ .824,  .89,  .729} 22.39\% & \cellcolor[rgb]{ 1,  .949,  .8} 22.86\% & 6.67\% & \cellcolor[rgb]{ 1,  .988,  .949} 10.55\% & 0.00\% & \cellcolor[rgb]{ 1,  .953,  .816} 24.64\% \\
    \textbf{instructBLIP-vicuna-13B} & \cellcolor[rgb]{ 1,  .996,  .976} 10.91\% & \cellcolor[rgb]{ 1,  .984,  .929} 13.68\% & 5.71\% & \cellcolor[rgb]{ 1,  .953,  .808} 19.40\% & \cellcolor[rgb]{ 1,  .973,  .878} 15.71\% & \cellcolor[rgb]{ 1,  .976,  .902} 11.11\% & 6.25\% & \cellcolor[rgb]{ 1,  .992,  .969} 2.63\% & \cellcolor[rgb]{ 1,  .98,  .922} 18.84\% \\
    \textbf{instructBLIP-flan-t5-xl} & \cellcolor[rgb]{ 1,  .953,  .808} 20.30\% & \cellcolor[rgb]{ 1,  .953,  .804} 22.22\% & \cellcolor[rgb]{ 1,  .953,  .808} 20.00\% & \cellcolor[rgb]{ 1,  .957,  .824} 17.91\% & \cellcolor[rgb]{ 1,  .949,  .8} 22.86\% & \cellcolor[rgb]{ 1,  .965,  .851} 13.33\% & \cellcolor[rgb]{ 1,  .965,  .851} 18.36\% & \cellcolor[rgb]{ 1,  .949,  .8} 15.79\% & \cellcolor[rgb]{ .769,  .875,  .71} 33.33\% \\
    \textbf{instructBLIP-flan-t5-xxl} & \cellcolor[rgb]{ .98,  .945,  .792} 20.91\% & \cellcolor[rgb]{ .929,  .925,  .773} 24.36\% & \cellcolor[rgb]{ .843,  .898,  .741} 22.86\% & \cellcolor[rgb]{ .941,  .929,  .776} 20.90\% & \cellcolor[rgb]{ .816,  .89,  .729} 25.71\% & \cellcolor[rgb]{ .835,  .898,  .737} 20.00\% & \cellcolor[rgb]{ 1,  .953,  .816} 21.09\% & \cellcolor[rgb]{ 1,  .957,  .82} 14.47\% & \cellcolor[rgb]{ 1,  .969,  .871} 21.74\% \\
    \textbf{LLAVANEXT-7B-vicuna} & \cellcolor[rgb]{ .565,  .808,  .631} 26.67\% & \cellcolor[rgb]{ .976,  .941,  .792} 23.08\% & \cellcolor[rgb]{ .592,  .816,  .643} 26.67\% & \cellcolor[rgb]{ .941,  .929,  .776} 20.90\% & \cellcolor[rgb]{ .722,  .859,  .69} 27.14\% & \cellcolor[rgb]{ .341,  .733,  .541} 33.33\% & \cellcolor[rgb]{ .686,  .847,  .678} 26.56\% & \cellcolor[rgb]{ .804,  .886,  .725} 19.74\% & \cellcolor[rgb]{ .855,  .902,  .745} 30.43\% \\
    \textbf{LLAVANEXT-13B-vicuna} & \cellcolor[rgb]{ .675,  .843,  .675} 25.15\% & \cellcolor[rgb]{ .992,  .949,  .8} 22.65\% & \cellcolor[rgb]{ .78,  .878,  .714} 23.81\% & \cellcolor[rgb]{ .941,  .929,  .776} 20.90\% & \cellcolor[rgb]{ .722,  .859,  .69} 27.14\% & \cellcolor[rgb]{ .588,  .816,  .639} 26.67\% & \cellcolor[rgb]{ .831,  .894,  .733} 24.61\% & \cellcolor[rgb]{ 1,  .949,  .8} 15.79\% & \cellcolor[rgb]{ .937,  .929,  .776} 27.54\% \\
    \textbf{LLAVANEXT-34B-NH} & \cellcolor[rgb]{ .498,  .784,  .604} 27.58\% & \cellcolor[rgb]{ .361,  .741,  .549} 39.32\% & \cellcolor[rgb]{ .659,  .839,  .667} 25.71\% & \cellcolor[rgb]{ .341,  .733,  .541} 28.36\% & \cellcolor[rgb]{ .341,  .733,  .541} 32.86\% & \cellcolor[rgb]{ .588,  .816,  .639} 26.67\% & \cellcolor[rgb]{ .373,  .745,  .553} 30.86\% & \cellcolor[rgb]{ .737,  .867,  .698} 21.05\% & \cellcolor[rgb]{ .384,  .749,  .561} 46.37\% \\
    \bottomrule
    \end{tabular}%
  }
    \caption{\model evaluation results on various multimodal LLMs across each multi-modal capability. Accuracy results are presented as  $\%$, with a maximum possible accuracy of $100\%$. Models achieving the highest scores are highlighted \textcolor{green}{\textbf{green}}, while lower-scoring models are highlighted \textcolor{olive}{\textbf{yellow}}.}
    \label{tab:capability}%
\end{table*}%

\subsection{Relationship between Model Size and Performance}

\begin{figure*}[htbp]
\centering  \includegraphics[width=0.8\linewidth]{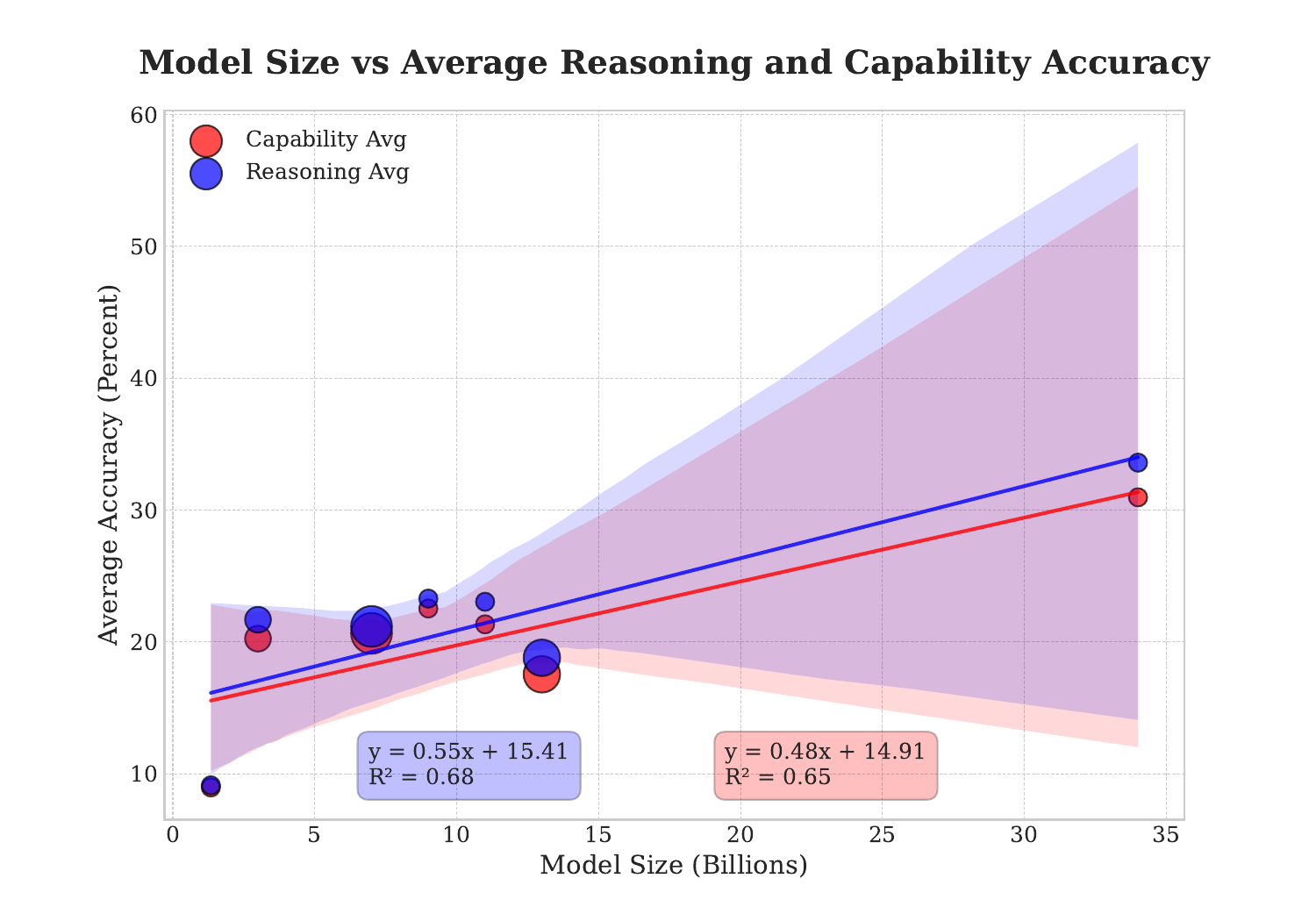}
  \caption{correlation between model size and average accuracy. The scatter plot uses varying dot sizes to represent the density of models with identical sizes.}
  \label{fig:size_v_acc}
\end{figure*}

Figure \ref{fig:size_v_acc} presents a comparative analysis of the model size and the average score achieved across all logical reasoning tasks and capabilities. Each plot includes a shaded region denoting the 95\% confidence interval for the regression estimate, visually representing the uncertainty associated with the regression line. Dot sizes in the scatter plot indicate the number of models with identical parameter counts, illustrating the distribution density. This visual evidence strongly suggests a positive correlation between larger model sizes and improved performance in \modelNoSpace. Specifically, as model size increases, performance tends to improve, indicating that larger models may have greater capacity to handle complex patterns and reasoning tasks.

\section{Conclusion} \label{conclude}
Reasoning skills are critical for solving complex tasks and serve as the foundation for many challenges that humans expect AI agents to tackle. However, the exploration of reasoning abilities in multimodal LLM agents remains limited, with most benchmarks and training datasets predominantly focused on traditional computer vision tasks like recognition. For multimodal LLMs to excel in critical thinking and complex tasks, they must comprehend the underlying logical relationships inherent in these challenges.

To address this gap, we introduce \modelNoSpace, a novel benchmark designed to evaluate multimodal LLMs through a comprehensive assessment of logical reasoning capabilities. This benchmark features a dataset of \datacount samples covering five distinct reasoning skills, providing a robust platform for evaluating cutting-edge multimodal models. Our evaluation aims to shed light on the current state of logical reasoning in multimodal LLMs.

To facilitate straightforward evaluation, we employ an LLM-based multiple-choice question-answer extractor, which helps mitigate the non-deterministic nature often associated with multimodal LLM outputs. While \model primarily focuses on explicit logical reasoning tasks isolated from real-life contexts, this approach represents a crucial step toward understanding fundamental reasoning skills. However, it is equally important to explore how AI agents perform tasks that blend abstract reasoning with real-world scenarios, a direction that will guide our future research endeavors.

\newpage

\section*{Acknowledgements}
We extend our sincere appreciation to the student researchers at the University of California, Los Angeles, for their diligent efforts in the manual annotation and validation of our dataset: Evan Li, Srinath Saikrishnan, Lawrence Li, and Oscar Cooper Stern.

\bibliographystyle{unsrt}
\bibliography{main}

\appendix

\onecolumn
 \begin{center}
    {\Large \textbf{Appendix: LogicVista: Multimodal LLM Logical Reasoning Benchmark in Visual Contexts}}
 \end{center}

\section{Examples of \model Logical Reasoning Data} \label{app:log}

\begin{table}[htbp]
\scriptsize
\caption{Three samples requiring inductive logical reasoning skills.}
\begin{minipage}{\textwidth}
\centering  
\scalebox{0.88}{
\begin{tabular}{r p{11.5cm} }
\toprule
\textbf{(Case A)} \\
&  \includegraphics[width=11.5cm]{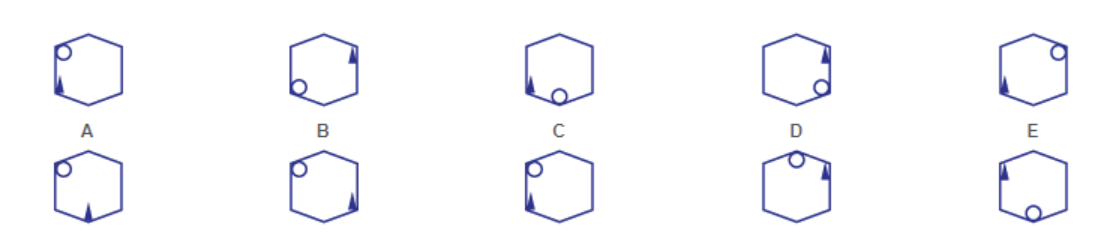} \\

\textbf{Q:} & Which choice (A, B, C, or D) completes the series? \\
\textbf{Answer:} & D \\
\textbf{Reasoning:} & In this example, there are two rules to be applied. The first is that the circle moves counter-clockwise in the hexagon. It follows that, in the following diagram, the circle will be in the upper corner of the hexagon, pointing to D as the answer. To confirm this, the second rule can be applied, according to which the position of the black triangle alternates between the bottom left and the top right. Thus, in the following diagram, the black triangle will need to be in the upper right corner of the hex. The answer is therefore definitely D.
\\
\textbf{Logical Reasoning Skill:} & Inductive \\ 
\textbf{Required capability} & Diagram \\ 
\midrule
\end{tabular}
}
\label{tab:samples_ind_a}
\end{minipage}
\vspace{-10pt}
\end{table}

\begin{table}[htbp]
\scriptsize
\caption{Three samples requiring inductive logical reasoning skills (Case B).}
\begin{minipage}{\textwidth}
\centering  
\scalebox{0.88}{
\begin{tabular}{r p{11.5cm} }
\toprule
\textbf{(Case B)} \\
~ & \includegraphics[width=11.5cm]{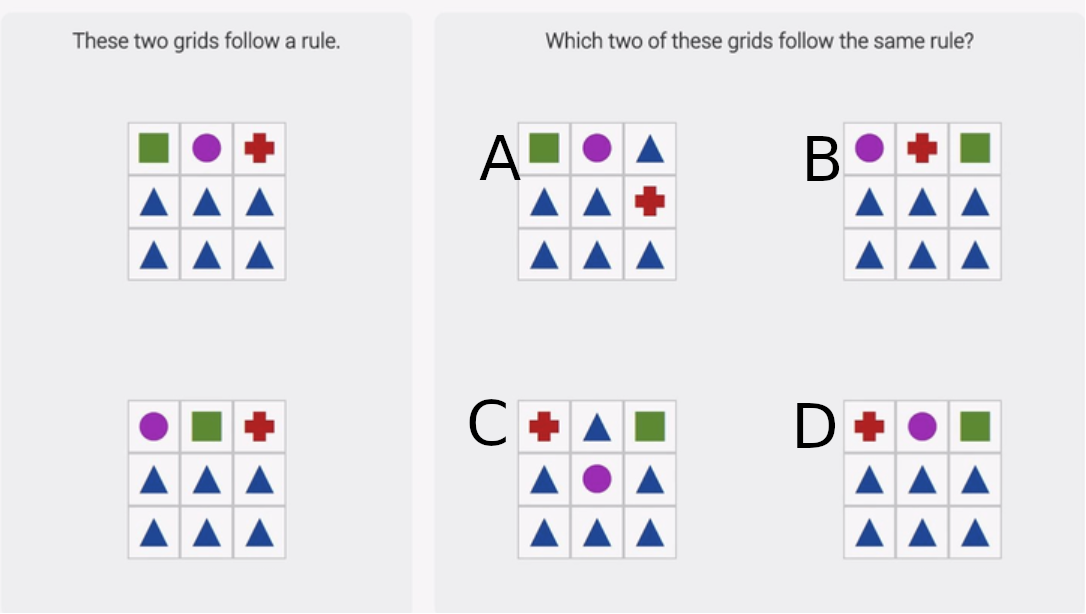} \\
\textbf{Q:} & Two grids containing colored symbols and following a common rule are presented. In the block on the right, four additional grids are presented. The candidate must find the two grids that follow the same rule out of these four options. What options (A, B, C, or D) follow this same rule? \\
\textbf{Answer:} & B, D \\
\textbf{Reasoning:} & In this example, it is easy to see that the rule governing the two grids on the left is: that blue triangles are present in each of the two bottom lines. This rule is followed in the two grids on the right. \\
\textbf{Logical Reasoning Skill:} & Inductive \\ 
\textbf{Required capability} & Diagram, OCR\\ 
\bottomrule
\end{tabular}
}
\label{tab:samples_ind_b}
\end{minipage}
\vspace{-10pt}
\end{table}

\begin{table}[htbp]
\scriptsize
\caption{Three samples requiring inductive logical reasoning skills (Case C).}
\begin{minipage}{\textwidth}
\centering  
\scalebox{0.88}{
\begin{tabular}{r p{11.5cm} }
\toprule
\textbf{(Case C)} \\
~ & \includegraphics[width=11.5cm]{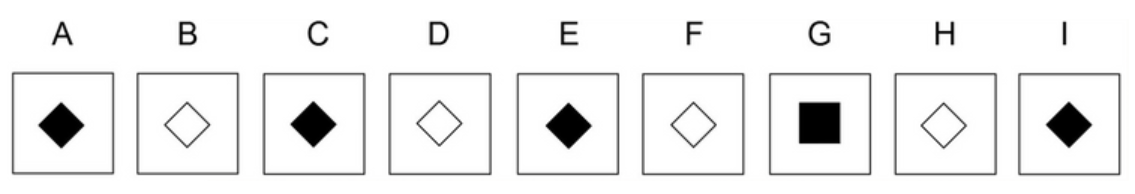} \\
\textbf{Q:} & Who is the odd-one-out? Select answers from A-I. \\
\textbf{Answer:} & G \\
\textbf{Reasoning:} & Element G constitutes the exception and is therefore the correct answer. \\
\textbf{Logical Reasoning Skill:} & Inductive \\ 
\textbf{Required capability} & Diagram \\ 
\midrule
\end{tabular}
}
\label{tab:samples_ind_c}
\end{minipage}
\vspace{-10pt}
\end{table}

\begin{table}
\caption{Three samples requiring deductive logical reasoning skills (Case A).}
\begin{minipage}{\textwidth}
\centering  
\scalebox{0.88}{
\begin{tabular}{r p{11.5cm} }
\toprule
\textbf{(Case A)} \\
&  \includegraphics[width=11.5cm]{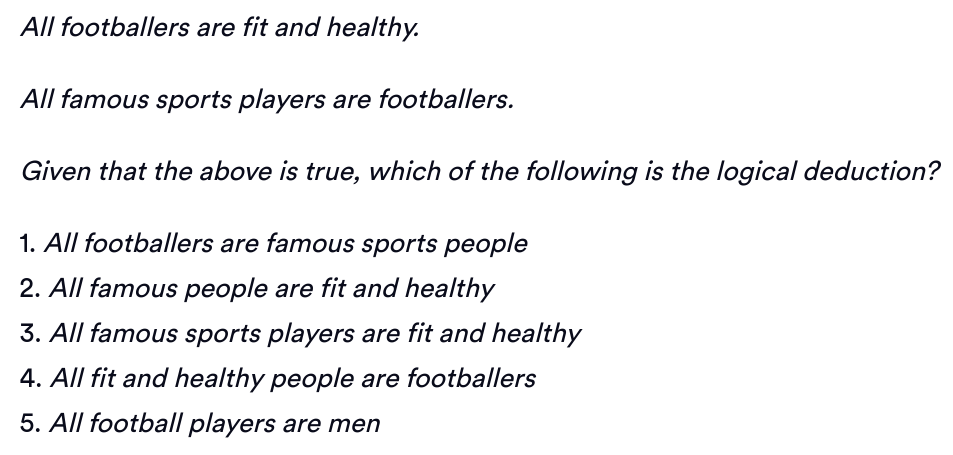} \\
\textbf{Q:} & Which is the correct answer according to the image? Select from 1-5? \\
\textbf{Answer:} & 3 \\
\textbf{Reasoning:} & Using deductive reasoning, the only logical answer is 3. To get to this answer, you need to simplify the given facts. All famous sports players are footballers, and all footballers are fit and healthy. We can not deduce that all footballers are famous sports people, as we have not got that information. We can not deduce that all famous people are fit and healthy, because the fact is about famous sports people. This is the logical answer. This information is not given; all footballers are fit and healthy but we can not logically link that all fit and healthy people are footballers. This is obviously incorrect, as gender is not mentioned at all in the question. \\
\textbf{Logical Reasoning Skill:} & Deductive \\ 
\textbf{Required capability:} & OCR \\ 
\midrule
\end{tabular}
}
\label{tab:samples_ded_a}
\end{minipage}
\end{table}

\begin{table}
\caption{Three samples requiring deductive logical reasoning skills (Case B).}
\begin{minipage}{\textwidth}
\centering  
\scalebox{0.88}{
\begin{tabular}{r p{11.5cm} }
\toprule
\textbf{(Case B)} \\
&  \includegraphics[width=11.5cm]{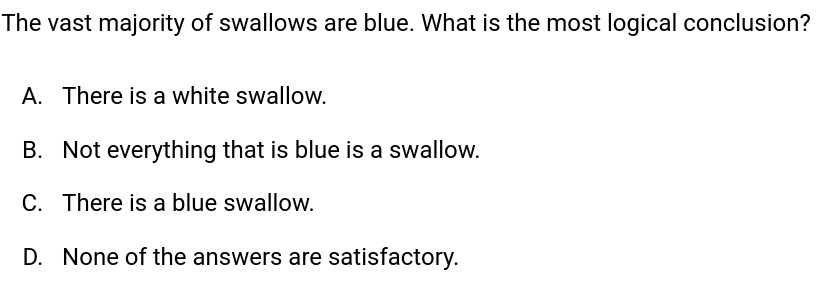} \\
\textbf{Q:} & What is the correct answer to the question in the image? Select from A-D. \\
\textbf{Answer:} & C \\
\textbf{Reasoning:} & The vast majority of swallows are blue so the answer must be C: there is a blue swallow. \\
\textbf{Logical Reasoning Skill:} & Deductive \\ 
\textbf{Required capability:} & OCR \\ 
\midrule
\end{tabular}
}
\label{tab:samples_ded_b}
\end{minipage}
\end{table}

\begin{table}
\caption{Three samples requiring deductive logical reasoning skills (Case C).}
\begin{minipage}{\textwidth}
\centering  
\scalebox{0.88}{
\begin{tabular}{r p{11.5cm} }
\toprule
\textbf{(Case C)} \\
&  \includegraphics[width=11.5cm]{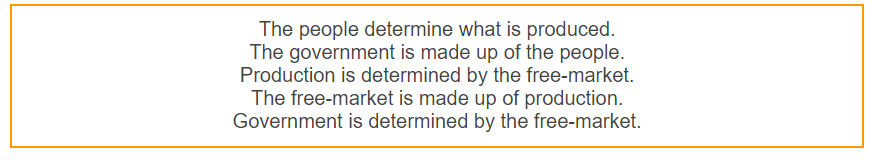} \\
\textbf{Q:} & What is produced is determined by the people. Select from A, B, and C. (A) True (B)False (C)Insufficient Information? \\
\textbf{Answer:} & A \\
\textbf{Reasoning:} & Line 1 states that the people determine what is produced. Line 2 states that the government is made up of the people. Therefore, the people determine what is produced. This is a syllogism. Thus, this statement is true. \\
\textbf{Logical Reasoning Skill:} & Deductive \\ 
\textbf{Required capability:} & OCR \\ 
\bottomrule
\end{tabular}
}
\label{tab:samples_ded_c}
\end{minipage}
\end{table}

\begin{table}
\scriptsize
\caption{Three samples requiring numerical logical reasoning skills (Case A).}
\begin{minipage}{\textwidth}
\centering  
\scalebox{0.88}{
\begin{tabular}{r p{11.5cm} }
\toprule
\textbf{(Case A)} \\
&  \includegraphics[width=6.5cm]{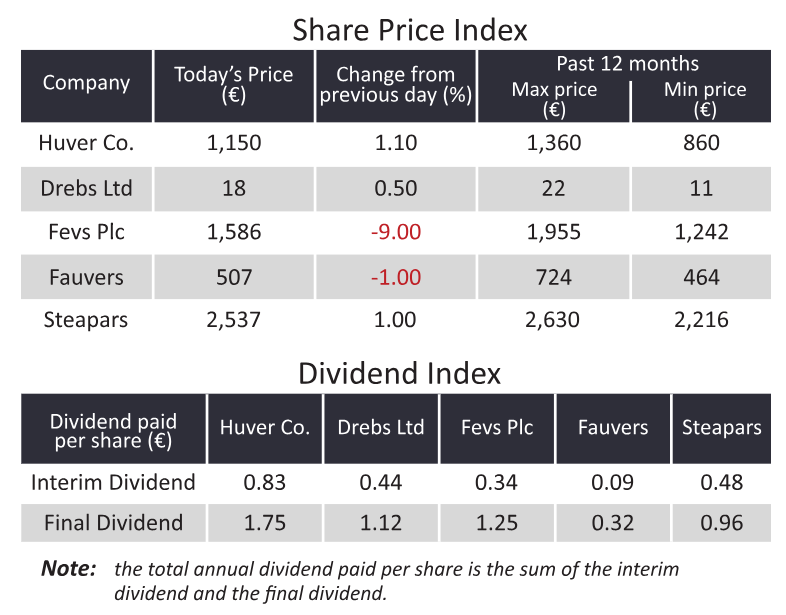} \\
\textbf{Q:} & Which share had the largest difference between the highest and lowest price over the last 12 months? Select from A, B, C, D and E. (A) Huver Co. (B) Drebs Ltd (C) Fevs Plc (D) Fauvers (E) Steapars \\
\textbf{Answer:} & C \\
\textbf{Reasoning:} & Step 1- Calculate the difference between the maximum and the minimum prices. Huver Co. = 1,360 - 860 = 500 Drebs Ltd = 22 - 11 = 11 Fevs Plc = 1,955 - 1,242 = 713 Fauvers = 724 - 464 = 260 Steapars = 2,630 - 2,216 = 414. Tip: Notice the wording of the question is asking for the share with the largest absolute change in price, NOT the largest percentage change, which would have been Drebs Ltd. If the question had wanted the percentage change it would have used the word percentage. Thus the correct answer is (C) Fevs Plc \\
\textbf{Logical Reasoning Skill:} & Numerical \\ 
\textbf{Required capability:} & OCR \\ 
\midrule
\end{tabular}
}
\label{tab:samples_num_a}
\end{minipage}
\end{table}

\begin{table}
\scriptsize
\caption{Three samples requiring numerical logical reasoning skills (Case B).}
\begin{minipage}{\textwidth}
\centering  
\scalebox{0.88}{
\begin{tabular}{r p{11.5cm} }
\toprule
\textbf{(Case B)} \\
&  \includegraphics[width=6.5cm]{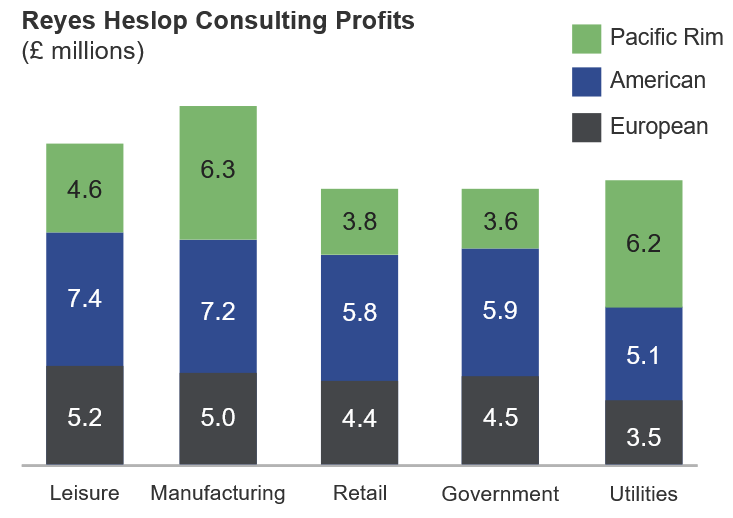} \\
\textbf{Q:} & Reyes Heslop had a target for Leisure profits to be a quarter of their total profits. Assuming profits in other areas remain the same, by how much did the Leisure profits miss this target? Select from A, B, C, D and E. (A) 31.8 million (B) 32.4 million (C) 32.7 million (D) 33.2 million (E) 33.4 million \\
\textbf{Answer:} & D \\
\textbf{Reasoning:} & Step 1- Calculate the total Reyes Heslop profits across all areas other than Leisure. (6.3 + 7.2 + 5.0) + (3.8 + 5.8 + 4.4) + (3.6 + 5.9 + 4.5) + (6.2 + 5.1 + 3.5) = 61.3 million. Step 2- This needs to be 1/4 of all profits for the condition to be met. Therefore all profits, across all sectors, would be 61.3 / 75\% = 81.7333 million. Step 3- Now we look at the difference between actual and target Leisure profits. Actual = (4.6 + 7.4 + 5.2) = 17.2 Target = (81.7333 - 61.3) = 20.4333 Shortfall = 3.2333 (millions) Thus the correct answer is (D) 33.2 million \\
\textbf{Logical Reasoning Skill:} & Numerical \\ 
\textbf{Required capability:} & Diagram, OCR \\ 
\midrule
\end{tabular}
}
\label{tab:samples_num_b}
\end{minipage}
\end{table}

\begin{table}
\scriptsize
\caption{Three samples requiring numerical logical reasoning skills (Case C).}
\begin{minipage}{\textwidth}
\centering  
\scalebox{0.88}{
\begin{tabular}{r p{11.5cm} }
\toprule
\textbf{(Case C)} \\
&  \includegraphics[width=6.5cm]{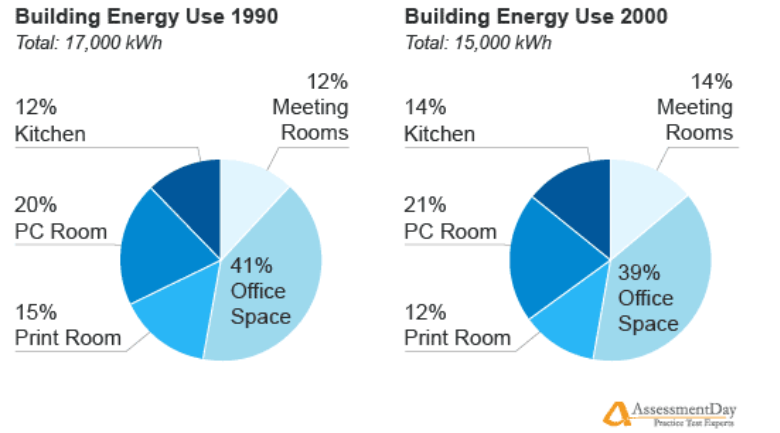} \\
\textbf{Q:} & Which space experienced the smallest reduction in kWh used between 1990 and 2000? Select from A, B, C, and D. (A) Office Space (B) Print Room (C) Meeting Rooms (D) PC Room \\
\textbf{Answer:} & D \\
\textbf{Reasoning:} & Step 1- Calculate the value of kWh for 1990 and 2000 for each of the rooms. Room 1990 per kWh 2000 per kWh Meeting Rooms 2.04 2.10 Office Space 6.97 5.85 Print Room 2.55 1.80 PC Room 3.40 3.15 Kitchen 2.04 2.10 Step 2- Subtract the kWh for 2000 from that of 1990 for each of the rooms. Room change (1990 - 2000) kWh Meeting Rooms -0.06 Office Space 1.12 Print Room 0.75 PC Room 0.25 Kitchen -0.06 Step 3- Look for the smallest positive value. Negative values represent an increase between 1990 and 2000. Tip- You only need to perform 4 calculations, as two of the rooms have the same values. Thus, the correct answer is (D) PC Room. \\
\textbf{Logical Reasoning Skill:} & Deductive \\ 
\textbf{Required capability:} & Diagram, OCR \\ 
\bottomrule
\end{tabular}
}
\label{tab:samples_num_c}
\end{minipage}
\end{table}

\begin{table}
\caption{Three samples requiring spatial logical reasoning skills (Case A).}
\begin{minipage}{\textwidth}
\centering  
\scalebox{0.88}{
\begin{tabular}{r p{11.5cm} }
\toprule
\textbf{(Case A)} \\
&  \includegraphics[width=6.5cm]{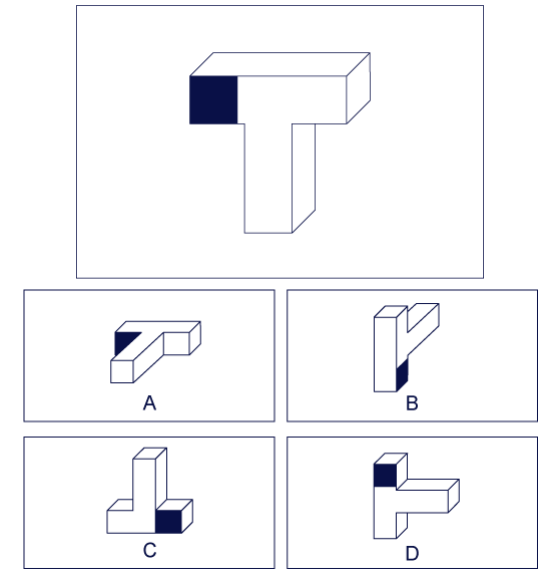} \\
\textbf{Q:} & Which figure is a rotation of the object? Select from A, B, C, and D. (A) (B) (C) (D) \\
\textbf{Answer:} & B \\
\textbf{Reasoning:} & The answer is B.  \\
\textbf{Logical Reasoning Skill:} & Spatial \\ 
\textbf{Required capability:} & Diagram \\ 
\midrule
\end{tabular}
}
\label{tab:samples_spat_a}
\end{minipage}
\end{table}

\begin{table}
\caption{Three samples requiring spatial logical reasoning skills (Case B).}
\begin{minipage}{\textwidth}
\centering  
\scalebox{0.88}{
\begin{tabular}{r p{11.5cm} }
\toprule
\textbf{(Case B)} \\
&  \includegraphics[width=4.5cm]{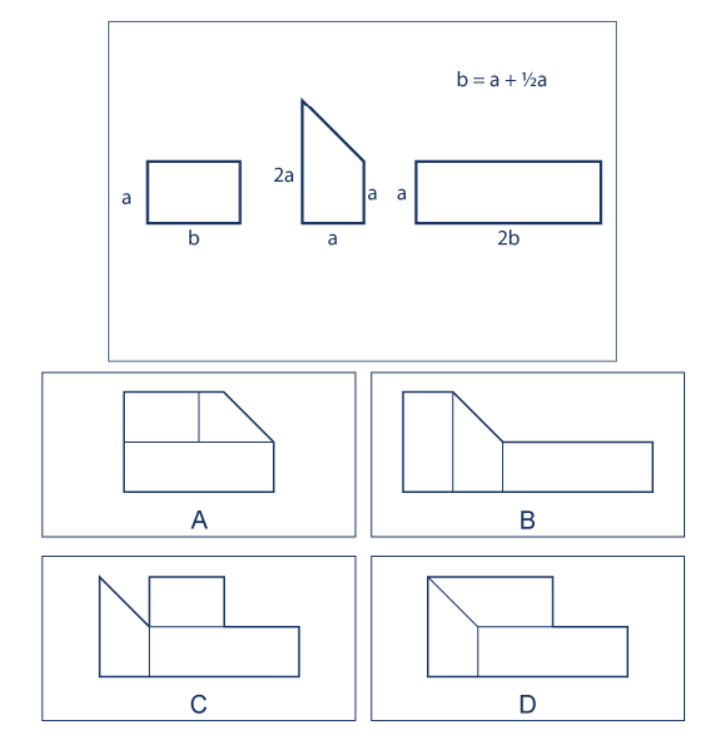} \\
\textbf{Q:} & Which figure can be formed with the given piece? Select from A, B, C, and D. (A) (B) (C) (D) \\
\textbf{Answer:} & C \\
\textbf{Reasoning:} & The answer is C.  \\
\textbf{Logical Reasoning Skill:} & Spatial \\ 
\textbf{Required capability:} & Diagram \\ 
\midrule
\end{tabular}
}
\label{tab:samples_spat_b}
\end{minipage}
\end{table}

\begin{table}
\caption{Three samples requiring spatial logical reasoning skills (Case C).}
\begin{minipage}{\textwidth}
\centering  
\scalebox{0.88}{
\begin{tabular}{r p{11.5cm} }
\toprule
\textbf{(Case C)} \\
&  \includegraphics[width=4.5cm]{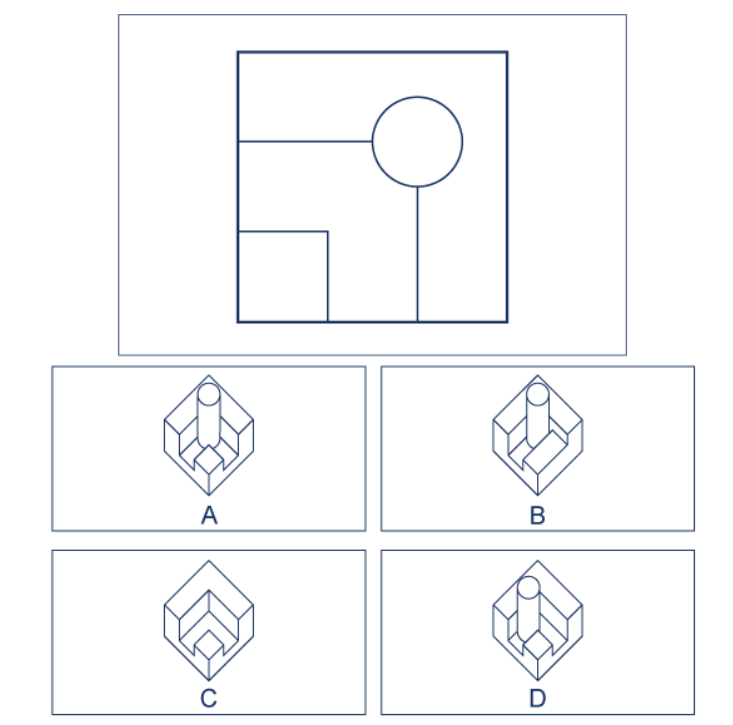} \\
\textbf{Q:} & To which object does the given top view correspond? Select from A, B, C, and D. (A) (B) (C) (D) \\
\textbf{Answer:} & A \\
\textbf{Reasoning:} & The answer is A. \\
\textbf{Logical Reasoning Skill:} & Spatial \\ 
\textbf{Required capability:} & Diagram \\ 
\bottomrule
\end{tabular}
}
\label{tab:samples_spat_c}
\end{minipage}
\end{table}

\begin{table}
\scriptsize
\caption{Three samples requiring mechanical logical reasoning skills (Case A).}
\begin{minipage}{\textwidth}
\centering  
\scalebox{0.88}{
\begin{tabular}{r p{11.5cm} }
\toprule
\textbf{(Case A)} \\
&  \includegraphics[width=6.5cm]{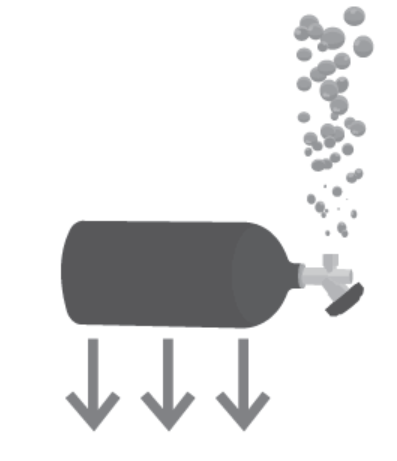} \\
\textbf{Q:} & A non-pressurised cylindrical metal tank filled with air is submerged underwater. As the air escapes, the tank gradually moves deeper underwater. Which statement provides the best reason for this motion? Select from A, B, C, D, and E. (A) The bubbles provide a downward thrust on the tank (B) The metal increases in density so it gets heavier (C) The bubbles lower the density of the water which lowers its buoyancy (D) Water replaces the air in the tank which makes it heavier (E) Impossible to tell \\
\textbf{Answer:} & D \\
\textbf{Reasoning:} & As air escapes the available space is quickly replaced with water, so the tank's density becomes the same as that of the water and with the added weight and density of the tank itself continues to sink. \\
\textbf{Logical Reasoning Skill:} & Mechanical \\ 
\textbf{Required capability:} & Diagram \\ 
\midrule
\end{tabular}
}
\label{tab:samples_mech_a}
\end{minipage}
\end{table}

\begin{table}
\scriptsize
\caption{Three samples requiring mechanical logical reasoning skills (Case B).}
\begin{minipage}{\textwidth}
\centering  
\scalebox{0.88}{
\begin{tabular}{r p{11.5cm} }
\toprule
\textbf{(Case B)} \\
&  \includegraphics[width=4.5cm]{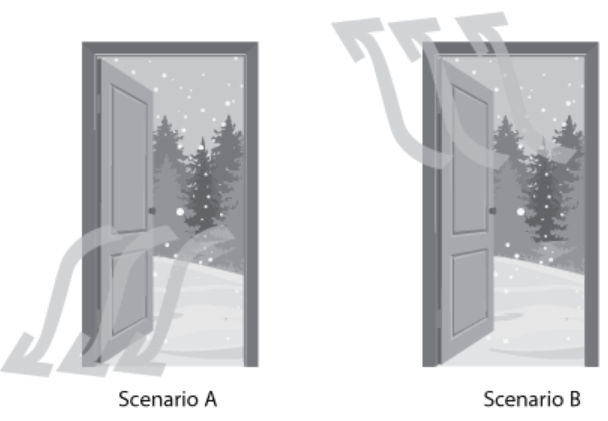} \\
\textbf{Q:} & It is a cold winter outside and a well-insulated house has its heater turned on. The front door is opened and cold air rushes in. If the wind speed outside is very low, how would the cold air enter the house? Select from A, B, C, D, and E. (A) Scenario A, the cold air will flow towards the floor (B) Scenario B, the cold air will flow towards the ceiling (C) A combination of A and B (D) The cold air will not enter the house (E) Impossible to tell \\
\textbf{Answer:} & A \\
\textbf{Reasoning:} & Cold air sinks, whereas hot air rises. The house and the air inside it are warmer than the outside air temperature, so if these two systems (house and outside) were to be suddenly connected (door opening) the cold air would sink and the hot air would sit above the cold air until the heat transferred between the two. \\
\textbf{Logical Reasoning Skill:} & Mechanical \\ 
\textbf{Required capability:} & Diagram \\ 
\midrule
\end{tabular}
}
\label{tab:samples_mech_b}
\end{minipage}
\end{table}

\begin{table}
\scriptsize
\caption{Three samples requiring mechanical logical reasoning skills (Case C).}
\begin{minipage}{\textwidth}
\centering  
\scalebox{0.88}{
\begin{tabular}{r p{11.5cm} }
\toprule
\textbf{(Case C)} \\
&  \includegraphics[width=4.5cm]{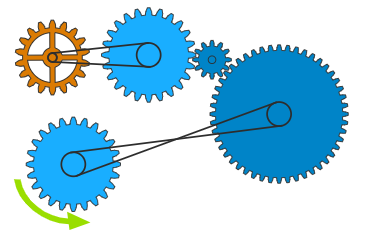} \\
\textbf{Q:} & In which direction does the orange gear rotate? Select from A, B, and C. (A) Clockwise (B) Counterclockwise (C) No rotation \\
\textbf{Answer:} & A \\
\textbf{Reasoning:} & The correct answer is clockwise. \\
\textbf{Logical Reasoning Skill:} & Mechanical \\ 
\textbf{Required capability:} & Diagram \\ 
\bottomrule
\end{tabular}
}
\label{tab:samples_mech_c}
\end{minipage}
\end{table}

\section{Examples of Different \model Capabilities Data} \label{app:cap}

\begin{table}[H]
\caption{Three samples of diagram, OCR, and mixed \model data (Case A).}
\begin{minipage}{\textwidth}
\centering  
\scalebox{0.88}{
\begin{tabular}{r p{11.5cm} }
\toprule
\textbf{(Case A)} \\
&  \includegraphics[width=4.5cm]{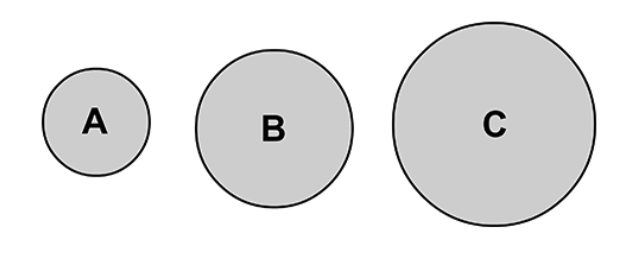} \\
\textbf{Q:} & Which ball is the heaviest? Select from A, B, C, and D. (A) A (B) B (C) C (D) CAN NOT SAY \\
\textbf{Answer:} & D \\
\textbf{Reasoning:} & The correct answer is D. \\
\textbf{Logical Reasoning Skill:} & Mechanical \\ 
\textbf{Required capability:} & Diagram \\ 
\midrule
\end{tabular}
}
\label{tab:samples_cap_a}
\end{minipage}
\end{table}

\begin{table}[H]
\caption{Three samples of diagram, OCR, and mixed \model data (Case B).}
\begin{minipage}{\textwidth}
\centering  
\scalebox{0.88}{
\begin{tabular}{r p{11.5cm} }
\toprule
\textbf{(Case B)} \\
&  \includegraphics[width=8.5cm]{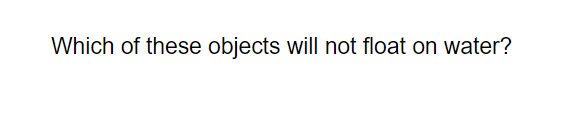} \\
\textbf{Q:} & Select from A, B, C, and D. (A) banana (B) scissors (C) empty plastic soda bottle (D) wooden pencil \\
\textbf{Answer:} & B \\
\textbf{Reasoning:} & The correct answer is B because scissors have metal and are most likely to sink. \\
\textbf{Logical Reasoning Skill:} & Deductive \\ 
\textbf{Required capability:} & OCR \\ 
\midrule
\end{tabular}
}
\label{tab:samples_cap_b}
\end{minipage}
\end{table}

\begin{table}[H]
\caption{Three samples of diagram, OCR, and mixed \model data (Case C).}
\begin{minipage}{\textwidth}
\centering  
\scalebox{0.88}{
\begin{tabular}{r p{11.5cm} }
\toprule
\textbf{(Case C)} \\
&  \includegraphics[width=6.5cm]{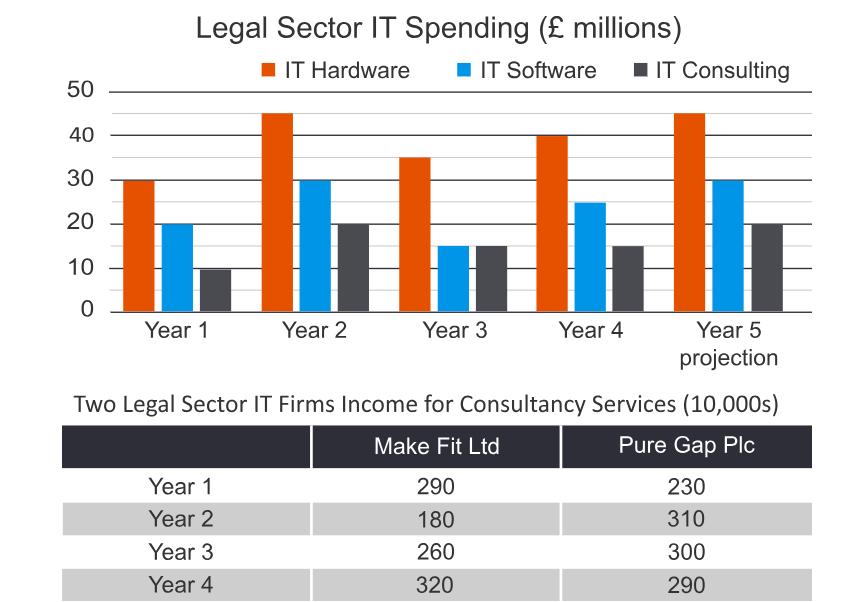} \\
\textbf{Q:} & Which of the following statements is false regarding legal sector spending between Year 4 and projected Year 5? Select from A, B, C, D, and E. (A) IT consulting will increase by 35 million. (B) IT consulting will match that of year 2. (C) IT software will exceed IT consulting. (D) Spending on IT hardware will decline. (E) None of these. \\
\textbf{Answer:} & D \\
\textbf{Reasoning:} & Step 1- Check in turn whether each statement is true or false: a) The projected spend on IT consulting is projected to increase by 35 million. Option A is true. b) The projected spend on IT consulting is 320 million, which matches year 2. Option B is true. c) The projected spend on IT software is 330 million and for IT consulting it is 320 million. Option C is true. d) There are increases projected for IT hardware, IT software, and consulting, therefore ``spending on IT hardware will decline'' is not true. The option for D is false. e) We see that option D is false, so E cannot be the correct answer. Thus the correct answer is (D) Spending on IT hardware, software, and consulting is projected to decline. \\
\textbf{Logical Reasoning Skill:} & Numerical \\ 
\textbf{Required capability:} & Diagram, OCR \\ 
\bottomrule
\end{tabular}
}
\label{tab:samples_cap_c}
\end{minipage}
\end{table}

\end{document}